\documentclass[10pt,twocolumn,letterpaper]{article}

\usepackage{cvpr}              

\usepackage{graphicx}
\usepackage{amsmath}
\usepackage{amssymb}
\usepackage{booktabs}
\usepackage{float} 
\usepackage{rotating}
\usepackage{stackengine}
\usepackage[normalem]{ulem}
\usepackage{array}
\usepackage{url}
\usepackage{lipsum,wrapfig}
\usepackage[accsupp]{axessibility}
\usepackage[utf8]{inputenc}
\newcommand{\Lidar}{LiDAR }
\newcommand{\Unet}{U-Net }

\newcolumntype{A}{>{\raggedleft\arraybackslash}m{0.6cm}}
\newcommand{\interval}[2]{\left[#1,#2\right]}
\makeatletter
\@namedef{ver@everyshi.sty}{}
\makeatother
\usepackage{tikz}
\usepackage[pagebackref,breaklinks,colorlinks]{hyperref}
\usepackage{multirow}
\usepackage{svg}
\usepackage[capitalize]{cleveref}
\crefname{section}{Sec.}{Secs.}
\Crefname{section}{Section}{Sections}
\Crefname{table}{Table}{Tables}
\crefname{table}{Tab.}{Tabs.}

\def\cvprPaperID{32} 
\def\confName{CVPR}
\def\confYear{2023}

\begin{document}

\title{DeepSim-Nets: Deep Similarity Networks for Stereo Image Matching}

\author{Mohamed Ali Chebbi\textsuperscript{1,2}
\and
Ewelina Rupnik\textsuperscript{2}
\and
Marc Pierrot-Deseilligny\textsuperscript{2} $\quad$ Paul Lopes\textsuperscript{1}
\and
{
	\textsuperscript{1}Thales, France \hspace{0.5cm}
 	\textsuperscript{2}Univ Gustave Eiffel, LASTIG, ENSG-IGN, F-94160 Saint-Mandé, France
}\\
{\tt \small \textcolor{magenta}{https://dalichebbi.github.io/DeepSimNets/}}
}
\maketitle
\begin{abstract}
We present three multi-scale similarity learning architectures, or DeepSim networks. These models learn pixel-level matching with a contrastive loss and are agnostic to the geometry of the considered scene. We establish a middle ground between hybrid and end-to-end approaches by learning to densely allocate all corresponding pixels of an epipolar pair at once. Our features are learnt on large image tiles to be expressive and capture the scene's wider context. We also demonstrate that curated sample mining can enhance the overall robustness of the predicted similarities and improve the performance on radiometrically homogeneous areas. We run experiments on aerial and satellite datasets. 
Our DeepSim-Nets outperform the baseline hybrid approaches and generalize better to unseen scene geometries than end-to-end methods. Our flexible architecture can be readily adopted in standard multi-resolution image matching pipelines. The code is available at \href{https://github.com/DaliCHEBBI/DeepSimNets}{https://github.com/DaliCHEBBI/DeepSimNets}.
\end{abstract}

\section{Introduction}
\label{sec:intro}
The availability of high quality large-scale stereo benchmark datasets \cite{wuteng,patil,dfc} prompted many neural network architectures for stereo matching. These architectures can be classified into two categories: hybrid and \textit{end-to-end}. To distinguish between matching and non-matching pixels, hybrid methods first extract features, then predict a similarity using a classifier (also referred to as \textit{similarity learning}). To infer the optimal surface, the known semi-global matching (SGM) follows~\cite{Hirsh}. Hybrid methods show good generalization properties to unseen scenes. However, they operate on small patches which imposes convolutional neural networks (CNN) with limited expressivity. 

End-to-end methods directly infer the surface from RGB images instead. They employ large image patches and deeper CNNs thus increase representations expressivity.  Most importantly, to leverage geometry and context-aware disparities, end-to-end methods combine texture cues from 2D feature representations with shape cues captured within 3D CNNs. Their disadvantage is that they rely on positive and fixed disparity range cost volumes. In real world scenarios, disparities can take any values, depending on the geometry of the scene and the camera acquisition geometry.\par
\begin{figure}[t!]
  \centering
   \includegraphics[width=0.99\linewidth]{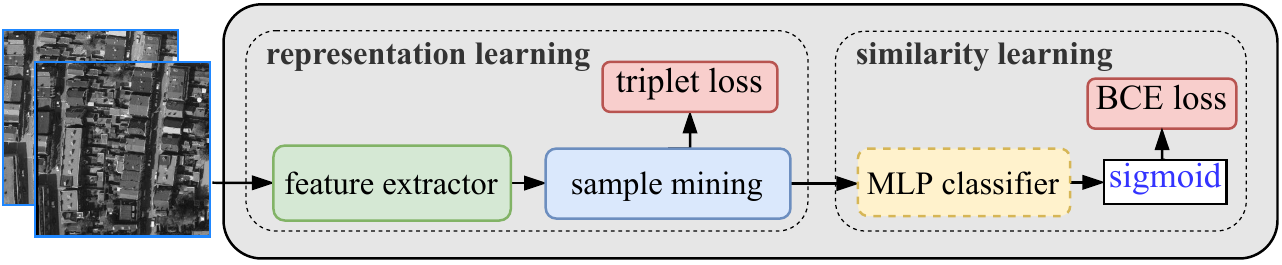}
   \vspace{-0.25cm}
   \caption{ {\textbf{DeepSim-Nets}}. The feature extractor is one of the three backbone variants: \Unet32, \Unet Attention, MS-AFF (see \cref{fig:msnet}). Reference feature, positive feature and negative feature sets are generated by the sample mining block and serve both representation and similarity learning tasks. The above architecture is a building block of the multi-resolution pipeline in \cref{fig:epipolarinference}.}
\label{fig:pipeline} 
\vspace{-0.5cm}
\end{figure}

\definecolor{mycyan}{rgb}{0.02,0.627,0.65}
\definecolor{mypurplee}{rgb}{1.0,0.0,1.0}
\begin{figure}[t!]
  \centering
   \subfloat[Left Epipolar]{\includegraphics[height=4cm,width=2.25cm]{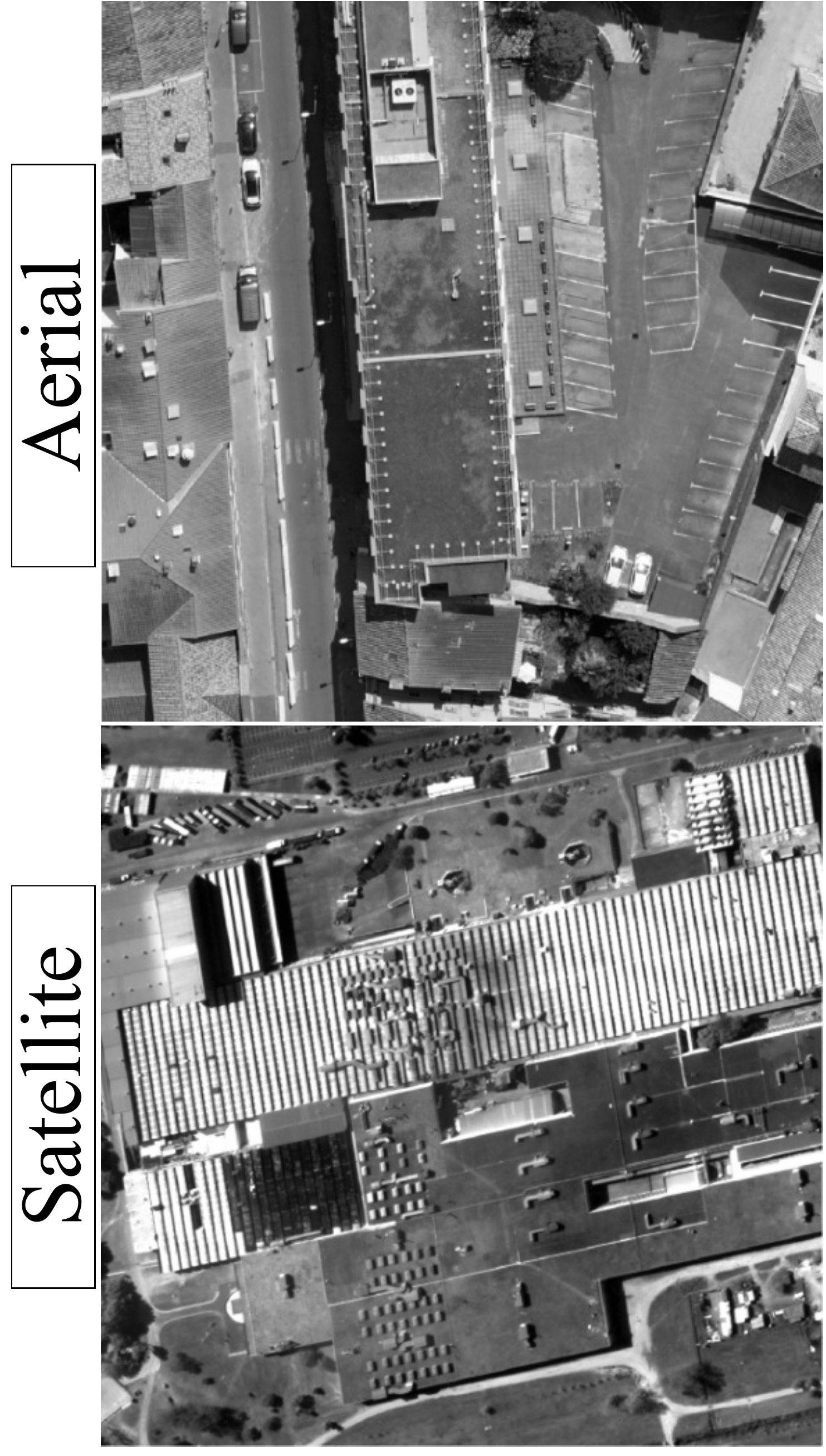}\label{fig:left}}
   \subfloat[\textbf{DeepSim-Net}]{\includegraphics[height=4cm,width=2cm]{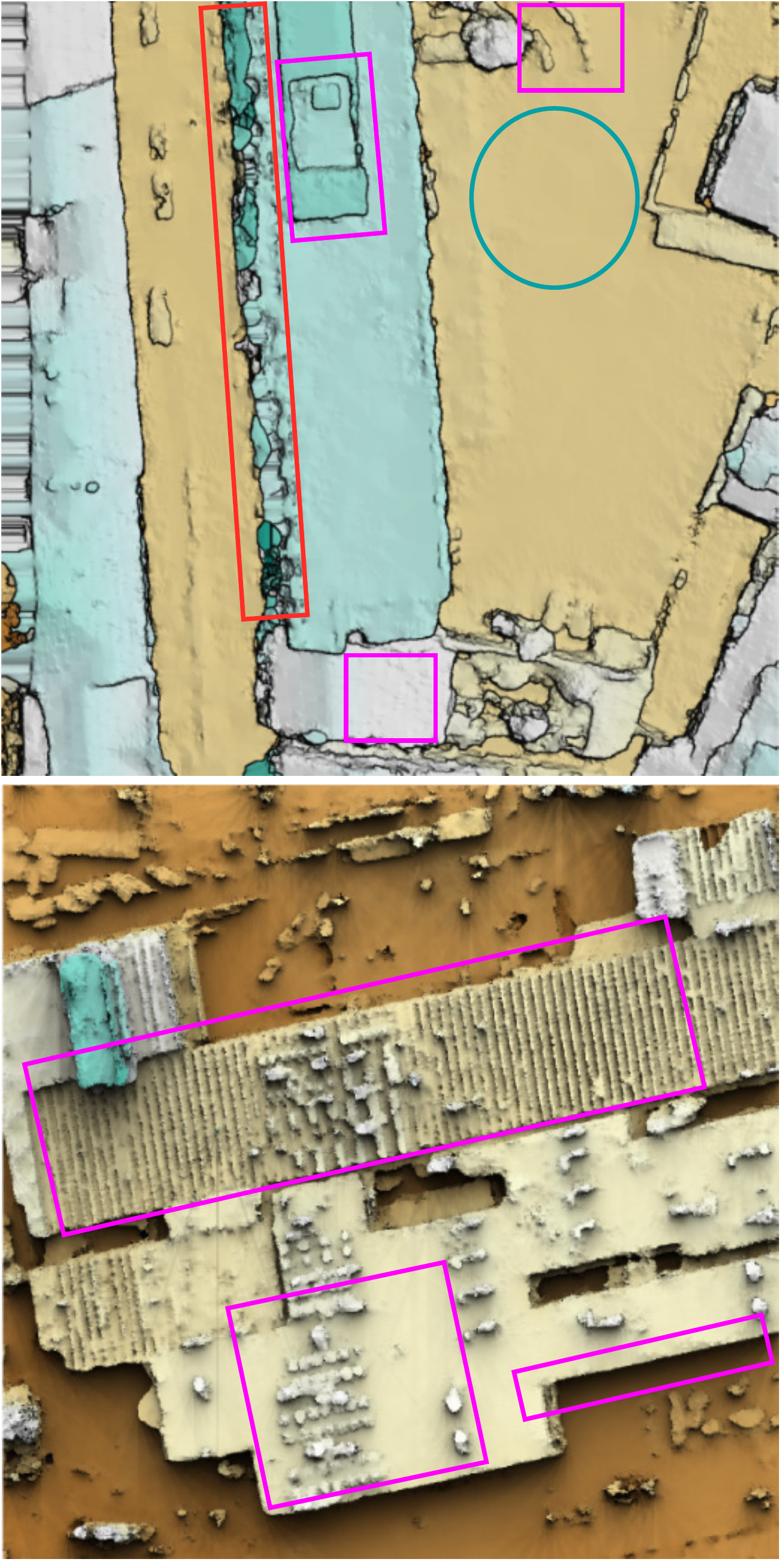}\label{fig:oursmsaff}}
   \subfloat[PSMNet]{\includegraphics[height=4cm,width=2cm]{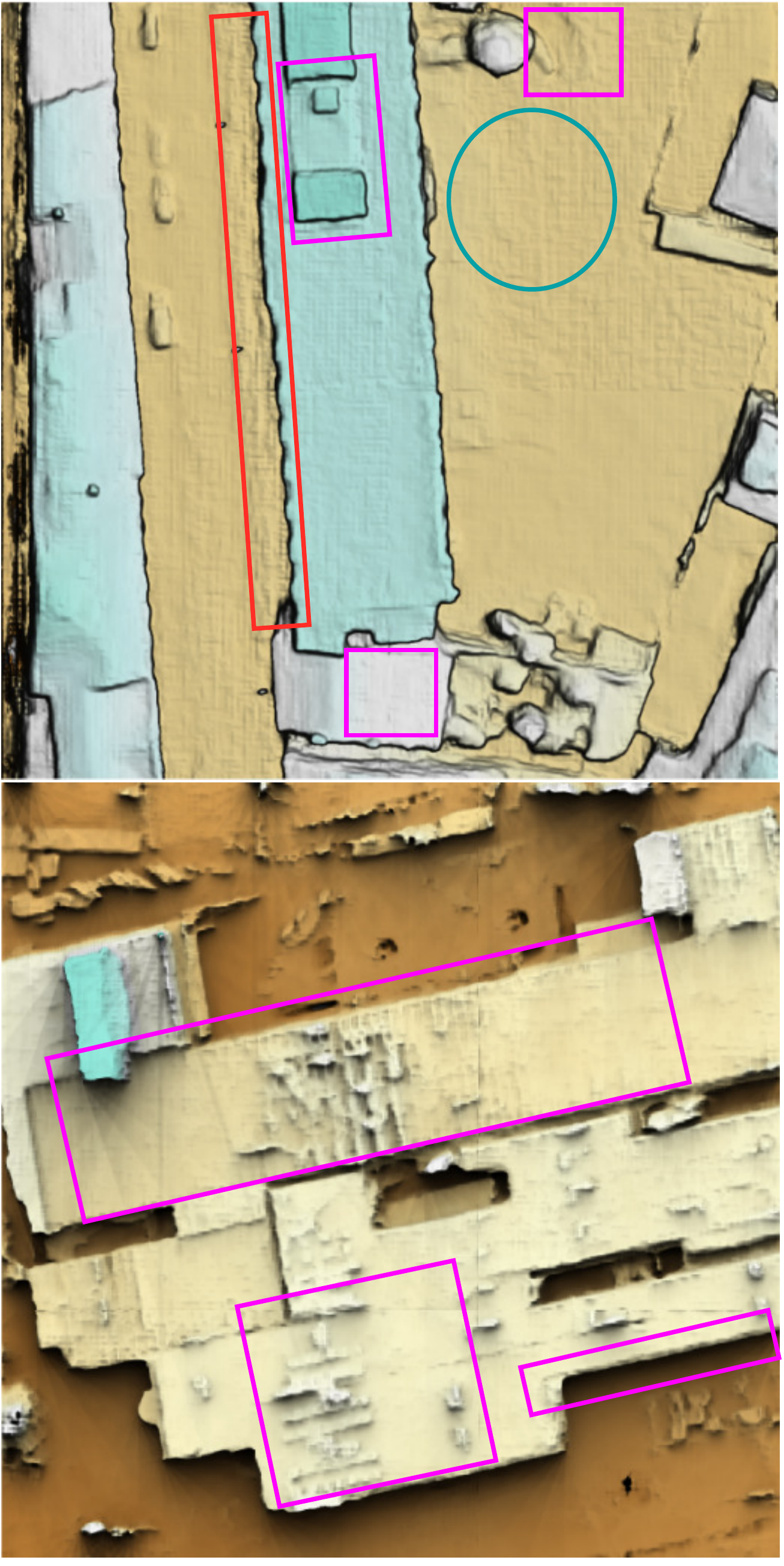}\label{fig:psmnetresults}}
   \subfloat[NCC]{\includegraphics[height=4cm,width=2cm]{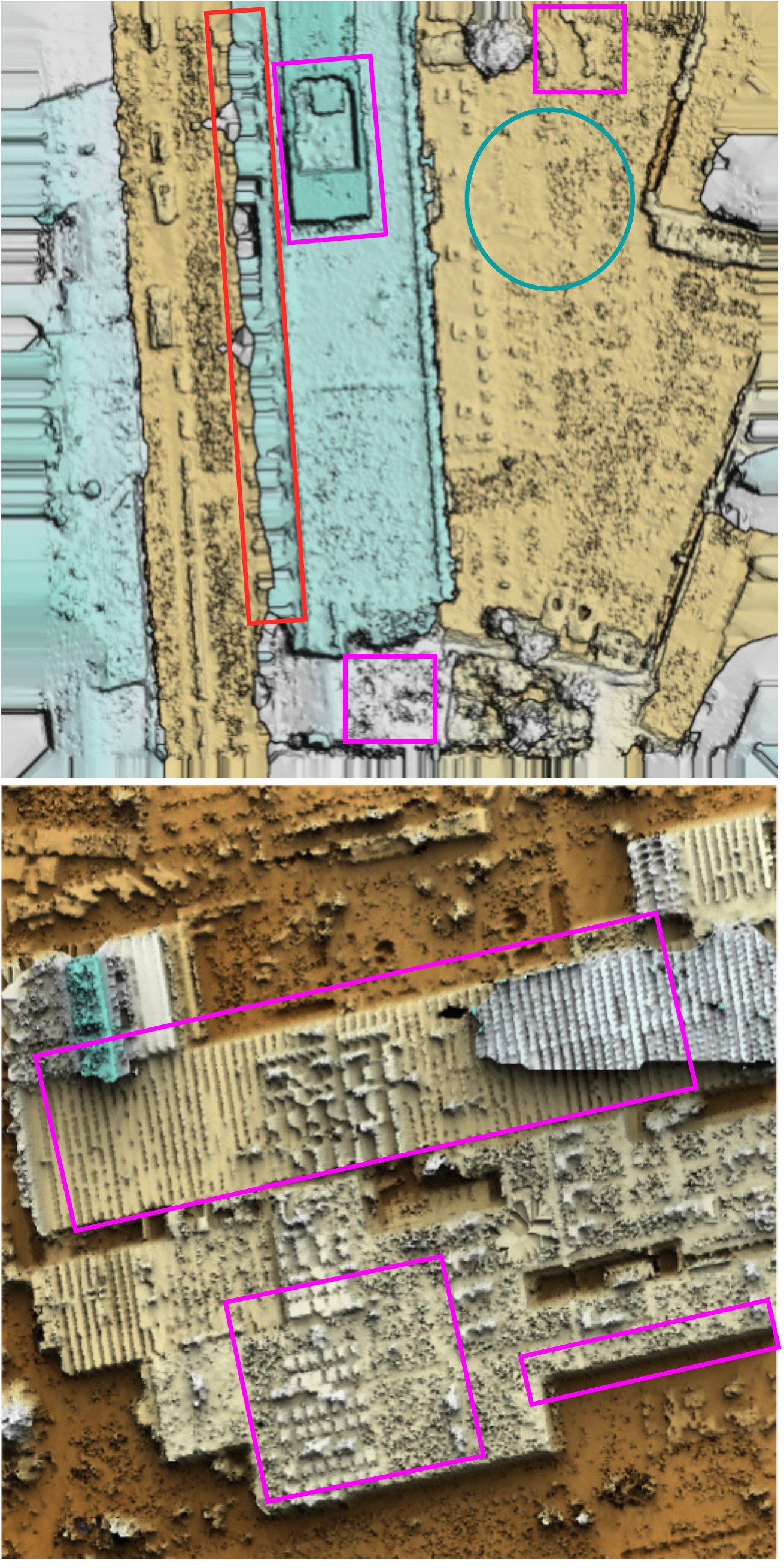}\label{fig:standard}}
   \vspace{-0.25cm}
   \caption{ {\textbf{Qualitative Results}}. We show two disparity maps generated from \emph{unseen} aerial (6cm) and satellite (WV-3, 30cm) stereo-pairs. On satellite data, our DeepSim-Net~\subref{fig:oursmsaff} performs best, while on aerial data the end-to-end PSMNet~\cite{PSMNet}~\subref{fig:psmnetresults} is best. The normalized cross-correlation (NCC)~\cite{micmac}~\subref{fig:standard} underperforms in both scenarios. On planar surfaces (\textcolor{mycyan}{$\bigcirc$}) DeepSim-Net yields faithful reconstructions, whereas PSMNet adds residual artefacts. On aerial data, PSMNet learns to interpolate in occluded areas \textcolor{red}{$\square$}, yet, it suppresses high-frequency details on satellite data and mis-constructs buildings' edges. Our DeepSim-Net recovers both buildings boundaries and fine details \textcolor{mypurplee}{$\square$}.} 
   \label{fig:teaser} 
   \vspace{-0.5cm}
\end{figure}
In this paper, we {revisit} the self-supervised deep similarity learning approach. To address the fixed disparity range flaw of end-to-end methods, we decouple similarity learning from surface inference, thus our method is hybrid (see \cref{fig:pipeline}). To enhance the expressivity of our features we no longer consider the local neighborhood of a pixel (small patch) but use contextually richer epipolar image pairs as input, see \cref{fig:teaser}. To our knowledge, the concept of deep similarity for stereo matching has not been introduced so far. Although counterintuitive, we show that an {\textit{off-the-shelf}} segmentation network such as \Unet \cite{unet,att_unet} can be trained to learn similarity semantics, provided a proper sample mining scheme {is adopted}. Finally, to reduce the network size we propose a hard-coded multi-scale feature extractor (see \cref{fig:msnet}) where specific non-weight sharing submodules learn specific scale cues. Unlike \Unet skip connection aggregation schemes, we employ an iterative attentional feature pooling mechanism to combine multi-scale features. Hence, we investigate the potential of implicit (U-Net) and explicit (ours) multi-scale learning. Note that multi-scale and multi-resolution are equivalent terms and we use them interchangeably throughout the paper.

To summarize, our main contributions are: 
(i) a new deep similarity learning architecture for stereo matching including a lightweight deep CNN architecture for feature learning that leverages hard-coded multi-scale features; (ii) a curated sample mining scheme to enable training deep architectures for our specified task; and
(iii) a hybrid cooperative pipeline that benefits from the robustness of hand-crafted similarity measures for lower resolutions and rich semantics features for higher resolutions.
%
\section{Related Works}
 \paragraph{Similarity learning.} Similarity learning focuses on predicting pixels' resemblance and leaves the spatial aggregation on the cost volume to standard SGM \cite{Hirsh,micmac} or global optimization \cite{CoxRoy}. During training, matching and non-matching patches extracted from epipolar images are fed to a CNN in a self-supervised fashion \cite{zbontar2016,visualembedding} or in a fully supervised fashion \cite{Han2015,Zagoruyko2015}. The task can be to either learn embeddings \cite{visualembedding} or the matching metric \cite{zbontar2016,Zagoruyko2015} or both \cite{Han2015}. In \cite{visualembedding} the authors propose a two-scale CNN architecture that endows features with robustness leveraged at different scales. MC-CNN \cite{zbontar2016} is the baseline for stereo-matching contrastive learning and addresses both embedding and similarity learning. 
Match-Net \cite{Han2015} adds more context by using $64 \times 64$ patches albeit a single descriptor is extracted for the center pixel. Alternatively, multi-view patch features can be used to learn similarity for multi-view stereo~\cite{konrad2017}. Others perform random forest classification to fuse hand-crafted similarity filled cost volumes \cite{cbmv2018}. Note that because similarity learning is bound to small patches, thus has a restrained receptive field, it is more susceptible to matching ambiguities (i.e., henceforth referred to as \textit{locality constraint}). To reduce stereo correspondence ambiguity, the similarities are backed by a regularization scheme that enforces surface regularity. An optimal regularization algorithm sets per-pixel edge-aware penalties which preserves thin structures and buildings outlines in the disparity map \cite{Hirsh}. Among the major merits of this hybrid scheme is the similarity which remains unrelated to a specific matching geometry.
 \par 
 A different stream of work has been devoted to explicitly incorporating semantic information into the stereo matching task. Coupling 3D semantic segmentation with disparity estimation in multi-task learning can provide excellent results, especially on diachronic images~\cite{bgam2021}. However, coarse objects frontiers may lower the disparity map quality in a strong supervised setting. 
\vspace{-0.5cm}
\paragraph{End-to-end disparity learning.} Among the first fully end-to-end stereo matching architectures is FlowNet \cite{FlowNet}. To learn and predict the optical flow images are fed to a CNN either stacked on top of each other (e.g., FlowNetSimple) or considered independent and followed by a hard-coded correlation layer (e.g., FlowNetCorr). 
The more modern GC-Net \cite{gcnet}, DeepPruner \cite{deepPruner} and PSMNet \cite{PSMNet} generate per-tile feature maps using a siamese CNN. Cost volume is subsequently built. GC-Net introduced a differentiable ArgMin (i.e., Soft-Argmin) that allows to train their network end-to-end. PSMNet builds upon that and introduces a deep 3D convolution Hourglass module to regularize the cost volume. The novelty of DeepPruner \cite{deepPruner} is in exploiting the learnt representations to prune per-pixel disparity range, thus being able to serve real-time applications. Alternatively, SGM is revisited in GA-Nets\cite{ganet} to leverage a differentiable optimization loss, and learn pixel-specific cost function parameters. Semi-global and local guided aggregation (SGA, LGA) layers are combined to balance regularity with edge awareness respectively.\par
So far, similarity learning being almost always governed by the locality constraint is not competitive with end-to-end methods that leverage both geometry and context~\cite{gcnet}. 
%
\definecolor{myblue}{rgb}{0.164, 0.07, 1.0}
\definecolor{mypink}{rgb}{1.0,0.46,0.87}
\definecolor{mygreen}{rgb}{0.44,0.74,0.274}
\definecolor{myred}{rgb}{1.0,0.176,0.15}
\section{Method}\label{sec:methods}
\begin{figure}[t!]
  \hspace{-0.5cm}
  \centering
   \includegraphics[width=0.5\textwidth]{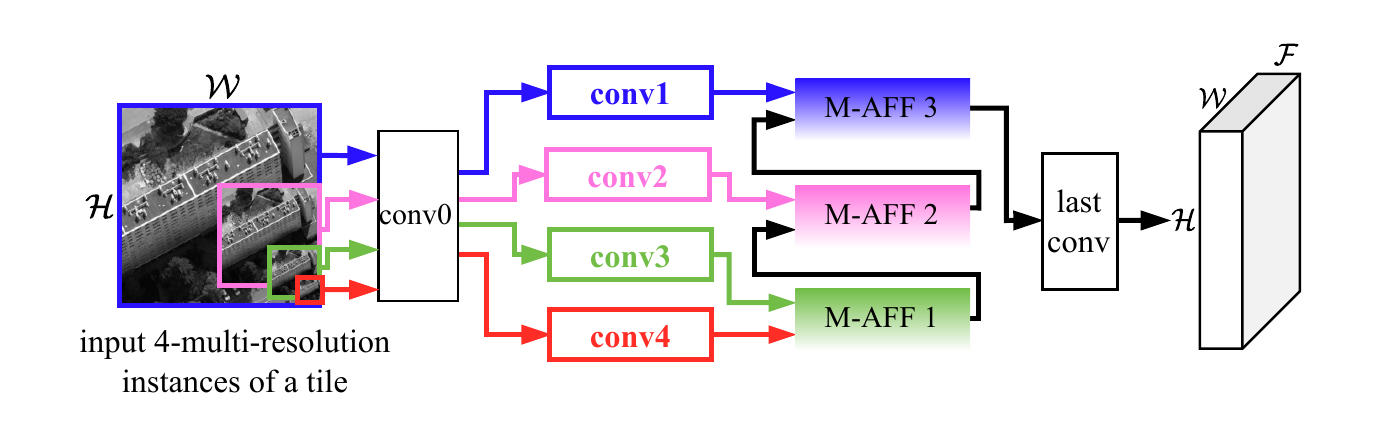}
   \vspace{-0.6cm}
   \caption{ \textbf{Our Lightweight Feature Extractor}. Explicit \textbf{M}utli-\textbf{S}cale self-\textbf{A}ttentional \textbf{F}eature learning and \textbf{F}usion (MS-AFF). conv0 is a CNN with 3 3x3 convolutional blocks. \textcolor{myblue}{\textbf{conv1}},\textcolor{mypink}{\textbf{conv2}},\textcolor{mygreen}{\textbf{conv3}} and \textcolor{myred}{\textbf{conv4}} are composed of 3 3x3 convolutional residual blocks~\cite{resnet}. They do not share weights and handle 4 different resolution feature maps extracted by conv0. The resulting embeddings are iteratively fused from lower to higher resolutions using stacked attentional fusion blocks M-AFF[1-3] (see \cref{fig:AFF}). The last\_conv consists of 3 3x3 convolutional blocks to produce $\mathcal{H}\times\mathcal{W}\times\mathcal{F}$ features.}
\label{fig:msnet}
\vspace{-0.5cm}
\end{figure}

We introduced \emph{DeepSim-Nets},{ a family of neural network architectures that learn to predict similarity score maps between tiles of pixels}. \cref{fig:pipeline} highlights \textit{DeepSim-Net}'s architecture {consisting} of a shared-backbone feature extractor, followed by a decision network that infers a similarity measure. The feature extractor consists of three variants:  (1)~\Unet32; (2) \Unet \textit{Attention} \cite{att_unet} performing gated attention feature pooling to aggregate encoder-decoder representations, and (3) our proposed explicit multi-scale feature learning module coupled with an adapted attentional pooling module \cite{aff} (see \cref{fig:msnet}).
Similarly to \cite{zbontar2016}, we follow the self-supervised learning paradigm. Traditionally, to retrieve context-aware and discriminative similarity measures, stereo correspondences were bound to a local-context support windows. However, locality leads to similarity ambiguity{, in particular} on textureless areas. 
To address this issue, we diverge from that idea. More specifically, non-local pixel embeddings that encapsulate similarity semantics are learnt by feeding large epipolar pairs $(768 \times 768)$  (i.e., \textit{tiles}) to the proposed large receptive field feature backbone variants. The tiles are randomly cropped from the  training stereo pairs.\par A suitable sample mining scheme that not only considers a single pair of patches \cite{zbontar2016,visualembedding,Han2015} but takes the whole set of features {at once} is then implemented. {Hence,} we allude to our sampling method as \textit{ensembling} and outline it in \cref{sec:pnsample}. 
{Finally,} we address the similarity learning problem as a classification task where both matching and non-matching feature pairs are encouraged to be apart from each other. The same {strategy is applied} to the distributions of matching and non-matching similarity metrics learnt by the subsequent Multi-Layer-Perceptron (MLP) classifier. 
%
%
\definecolor{myorange}{rgb}{0.84,0.71,0.34}
\subsection{Representation learning}
\label{sec:representation}
\begin{figure}[t]
    \centering
    \hspace*{1.4cm} \includegraphics[width=0.35\textwidth]{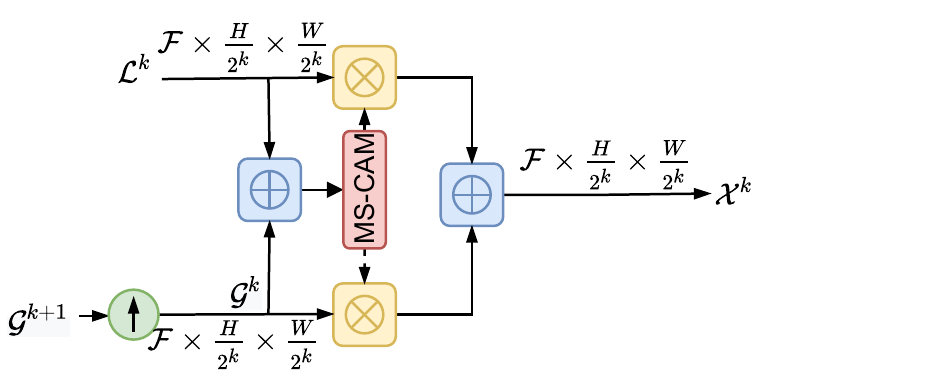}
    \vspace{-0.2cm}
    \caption{ \textbf{Attentional Multi-scale Feature Fusion} (M-AFF). It is a building block of our lightweight feature extractor MS-AFF in \cref{fig:msnet}. MS-CAM\cite{aff} learns to weight local and global embeddings contributions to the fused representations.\textcolor{blue}{$\oplus$} is the addition operator, \textcolor{myorange}{$\otimes$} is the Hadamard multiplication operator.}
    \label{fig:AFF}
    \vspace{-0.5cm}
\end{figure}
The feature extractor takes a grayscale \(\mathcal{H}\times \mathcal{W}\) image (i.e, tile) and outputs \(\mathcal{H} \times \mathcal{W}\times \mathcal{F}\) feature map  (see \cref{fig:pipeline}). We compute similarity scores along the epipolar line between a reference feature $f^{l}_{x,y} \in \mathbb{R}^\mathcal{F}$ from the left epipolar pair (i.e., left $(l)$ tile) and a set of possible features drawn from the right epipolar pair (i.e., right $(r)$ tile) $f^{r}_{x-i,y} \in \mathbb{R}^\mathcal{F}$ with $i \in [d_{min}, d_{max}]$ defining the disparity search space. $(x,y)$ are pixel locations. {The normalized dot product $\left< ., . \right>$ between a pair of features is inherently a similarity measure as it equals the cosine of the angle between them.} 
 Therefore, by training the backbone feature extractor to yield high similarity scores for matching feature pairs and low similarity scores for mismatching ones, the network learns robust and discriminative features {that encapsulate similarity cues}. It follows that for a set of reference features $\mathcal{X}$ drawn from the left tile, a set of matching features $\mathcal{X}_{+}$ and a set of non-matching features $\mathcal{X}_{-}$, all generated from the right tile, our triplet loss is:
 \begingroup
\setlength{\abovedisplayskip}{3pt}
\setlength{\belowdisplayskip}{4pt}
\begin{equation}
\label{eq:tri}
 \mathcal{L}_{3}=
           \sum_{(i,j) \in \mathcal{X}} \mathcal{O}(\mathcal{S}^{i,j}_{-}- \mathcal{S}^{i,j}_{+}+ m,0),
\vspace{-0.2cm}
\end{equation}
\endgroup
where $(i,j)$ denote feature coordinates in the reference feature map, $\mathcal{S}^{i,j}_{-}=\left<\mathcal{X}^{i,j}_{-},\mathcal{X}^{i,j}\right>$, $\mathcal{S}^{i,j}_{+}=\left<\mathcal{X}^{i,j}_{+},\mathcal{X}^{i,j}\right>$ are cosine similarities between features; $m$ is the separation margin; and $\mathcal{O}$ is the element-wise max operator. {We set $m$ empirically to ${0.3}$ and keep it fixed for all experiments.}
\vspace{-0.5cm}
\paragraph{Attentional feature pooling. }{Combining multiple and complementary types of features to obtain semantically stronger representations has proven beneficial in many application domains \cite{att_unet,aff,robert2022dva}.}
{Similarly, two sets of spatially consistent feature maps computed at different image scales encapsulate complementary cues as their respective fictive receptive fields differ. This observation was used in the \Unet architecture \cite{unet} where low and high level features are concatenated via long skip connections. Here, rather than blindly concatenating the multi-scale features as does \Unet, we introduce an aggregation strategy through a Multi-Scale self-Attention Feature Fusion (MS-AFF). Thanks to this explicit fusion we reduce the number of parameters by a factor of 10.} \par 
MS-AFF's key idea is presented in \cref{fig:msnet}, where $\mathcal{L}^{k}$ is a feature map of shape $\frac{H}{2^{k}}\times\frac{W}{2^{k}}\times\mathcal{F}$ denoted as local feature map and $\mathcal{G}^{k+1}$ is a more global feature of shape $\frac{H}{2^{k+1}} \times \frac{W}{2^{k+1}}\times \mathcal{F}$. Based on the multi-scale attention feature fusion module (MS-CAM)~\cite{aff} denoted by $\mathcal{M}$, we refine the feature map $\mathcal{X}^k$ at scale  $\textit{k}$ using the formula:
\begingroup
\setlength{\abovedisplayskip}{5pt}
\setlength{\belowdisplayskip}{5pt}
\begin{equation}
    \mathcal{X}^{k} = \mathcal{M}(\mathcal{L}^{k} \oplus  \mathcal{G}^{k}) \otimes \mathcal{L}^{k} + (1-\mathcal{M}(\mathcal{L}^{k} \oplus  \mathcal{G}^{k})) \otimes \mathcal{G}^{k}, 
\end{equation}
\endgroup
where $\mathcal{G}^{k}$ is a one level up-scaled version of $\mathcal{G}^{k+1}$. By extending this fusion concept to pyramidal features maps, multiple attentional fusion blocks  can be stacked on top of each other moving from coarser low-level contextually rich features to fine-grained high resolution features (see \cref{fig:AFF}). At each step, the result of the last aggregation is considered as a global feature for the next fusion block.

\subsection{Similarity learning}
{Our goal is to learn a powerful similarity function that predicts the matching likelihood between two embeddings. We believe that to describe complex relationships between corresponding pixels, the baseline dot product is insufficient. To that end, we feed the learnt representations to an MLP module acting as the decision function. Supervision is accomplished by the binary cross entropy (BCE) loss \cite{zbontar2016}.} %
Following the previously introduced notation, given the triplet of feature sets $\mathcal{X}$,$\mathcal{X}_{+}$ and $\mathcal{X}_{-}$, we formulate the per-tile BCE loss $\mathcal{L_{BCE}}$ as:
\begingroup
\setlength{\abovedisplayskip}{5pt}
\setlength{\belowdisplayskip}{2pt}
\begin{equation}
\mathcal{L}_{BCE}=
-\hspace{-0.25cm}\sum_{(i,j) \in \mathcal{X}}\mathcal{Y}^{i,j}_{-} \log(1-\mathcal{S}^{i,j}_{-}) + \mathcal{Y}^{i,j}_{+} \log(\mathcal{S}^{i,j}_{+}),
\end{equation}
\endgroup
where $\mathcal{S}^{i,j}_{-}=\Phi(\mathcal{C}(\mathcal{X}^{i,j}_{-},\mathcal{X}^{i,j}))$, $\mathcal{S}^{i,j}_{+}=\Phi(\mathcal{C}(\mathcal{X}^{i,j}_{+},\mathcal{X}^{i,j}))$. $\mathcal{C}$ is the concatenation operator and $\Phi$ is the MLP that maps features from $\mathbb{R}^{\mathcal{F} \times 2}$  to $\mathbb{R}$; $\mathcal{Y}_{+}$ and $\mathcal{Y}_{-}$ are positive and negative sample definition masks, respectively. A sample definition mask is a binary mask that defines the matching features locations included in the loss computation. 
\subsection{Sample {mining}}
\label{sec:pnsample}
\paragraph{Ensembles approach.} {Our sampling technique is designed to operate at a tile level, ensuring that the features learned by the network are consistent across entire objects (e.g., buildings, roads, etc.). Moreover, our approach involves presenting an \textit{ensemble} of samples in a single gradient update, which not only adds more spatial context but also prevents overfitting. Specifically, in one gradient step a feature can appear as a positive match to one feature, and at the same time as a negative match to several other features (see \cref{fig:sampling}). Patch-based shallow networks cannot capture spatial relationships in large objects, unlike our method.} 
\vspace{-0.7cm}
\paragraph{Sample Mining.} 
The quality and density of the dataset can impact the sample mining. For instance, coarse optical-\Lidar registration may produce false pixel correspondences, which can confuse positive and negative examples. To address this, we sample positive pixel examples around ground truth back-projected \Lidar points and negative pixel examples slightly further away from the ground truth. Additionally, we densify the \Lidar ground truth data using Delaunay interpolation to match the image data density, preserving high frequency changes and occlusion constraints while being a purely geometric approach.
\begingroup
\setlength{\intextsep}{0pt}%
\begin{figure}[t!]
  \centering
  \subfloat[GT.]{\includegraphics[width=0.12\textwidth]{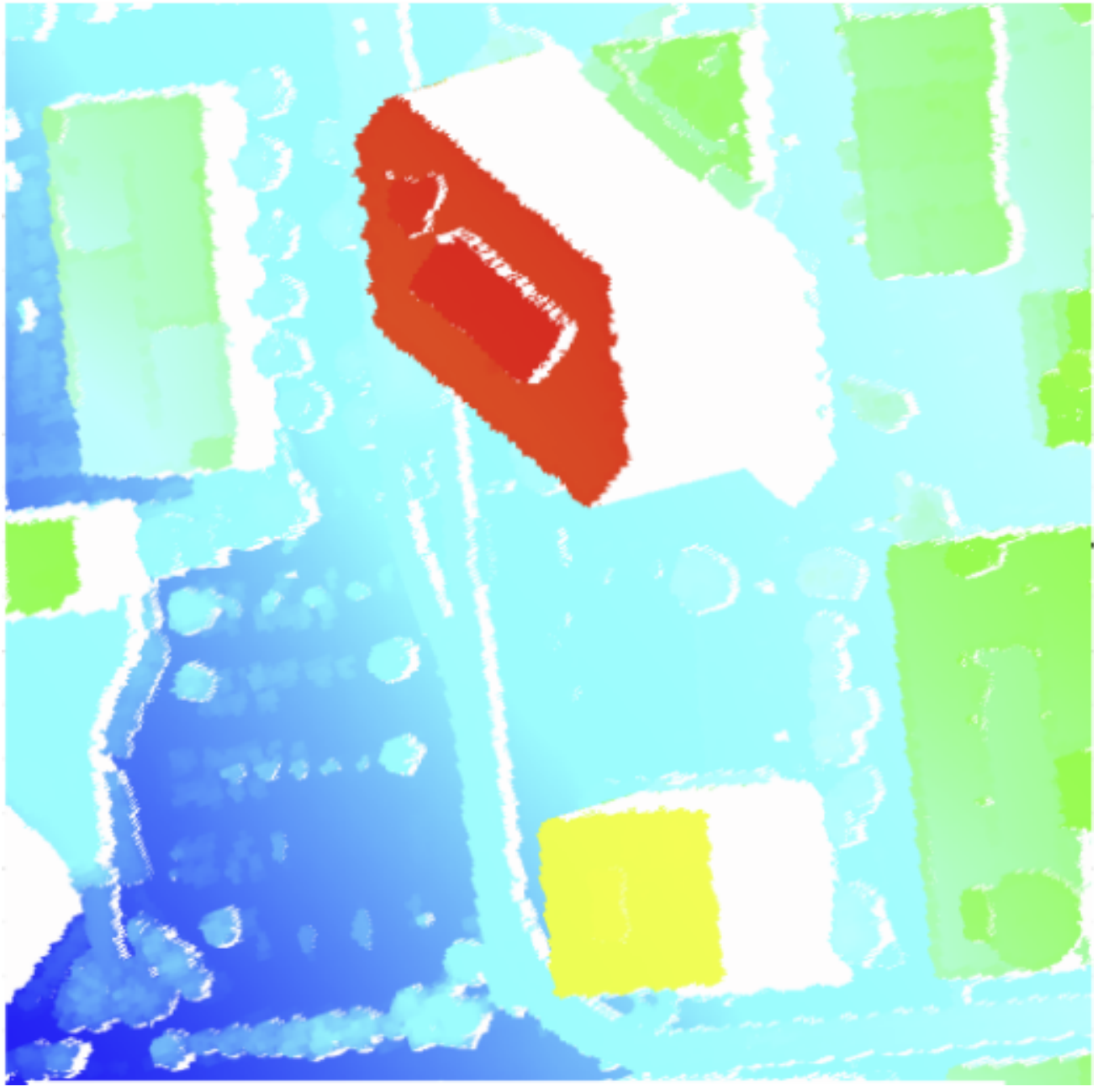}\label{fig:gt}}
  \hspace{0.1cm}
  \subfloat[Correspondences Sampling.]{\includegraphics[width=0.32\textwidth]{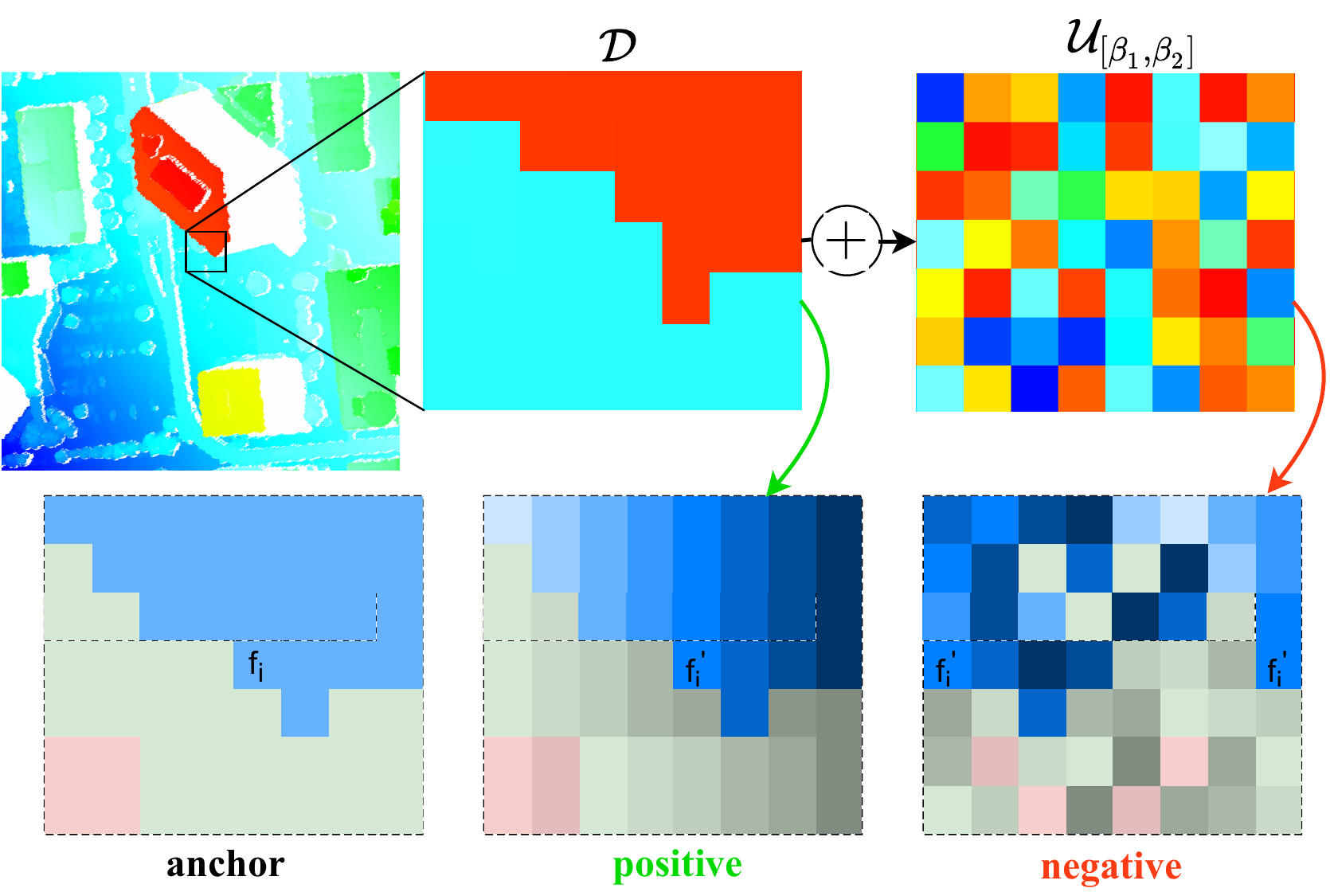}\label{fig:ideamatching}}\vspace{0.05cm}
  \subfloat[Triplet Feature Sets Generation.]{\includegraphics[width=0.44\textwidth]{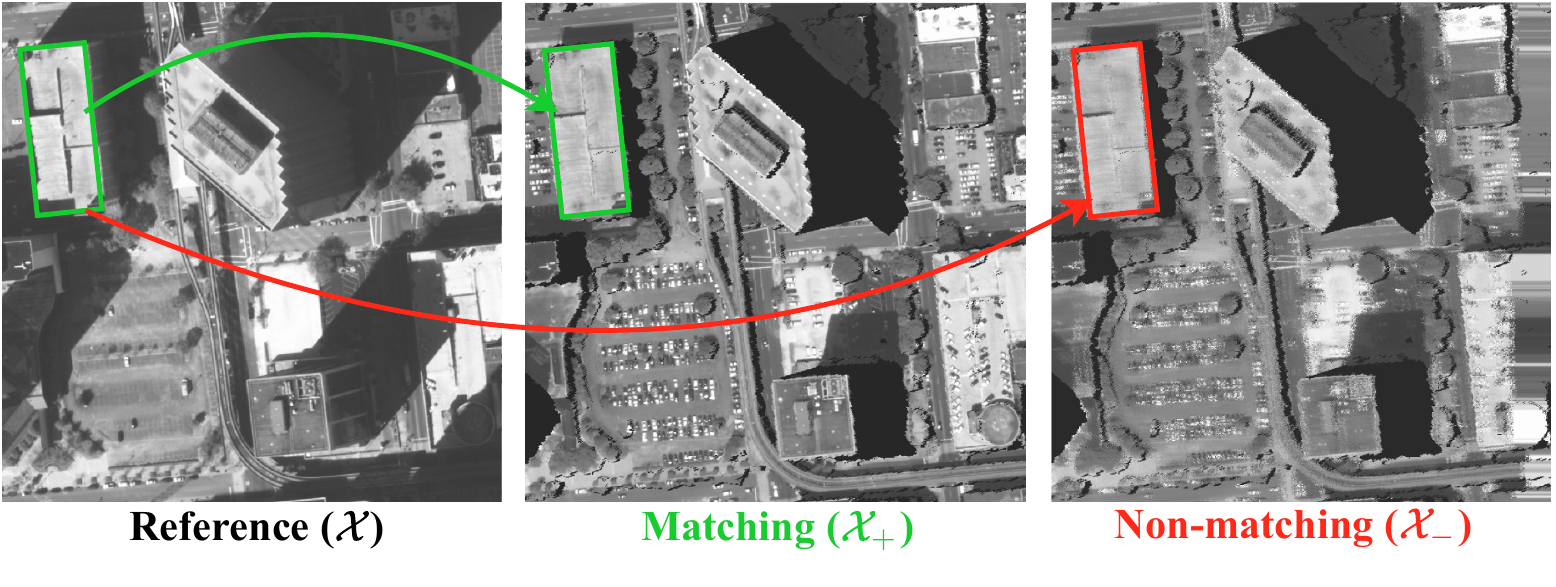}\label{fig:warping}}
   \vspace{-0.2cm}
   \caption{{\textbf{Ensembles Sample Mining Toy Example}}. \protect\subref{fig:gt} Ground truth (GT) disparities define matching pixels mappings ($\mathcal{D}$).  \protect\subref{fig:ideamatching} Let $(f_{i},f^{'}_{i})$ be a feature pair labelled as positive. Negative feature pair is picked randomly along the epipolar line in the vicinity of $f^{'}_{i}$. Note that $f^{'}_{i}$ {can have} both \textit{matching} (\textcolor{green}{positive}) and \textit{non-matching} (\textcolor{red}{negative}) states in the \textbf{same} gradient update. This enforces the matching uniqueness constraint while preventing overfitting. \protect\subref{fig:warping}~At tile level, the feature map is warped based on ground truth disparities (\textcolor{green}{$\large\rightarrow$}), and denoted as the matching feature map (\textcolor{green}{$\mathcal{X}_{+}$}). The non-matching feature map (\textcolor{red}{$\mathcal{X}_{-}$}) is obtained by warping the  feature (\textcolor{red}{$\large\rightarrow$}) map using random disparity offsets ($\mathcal{D}+\mathcal{U}_{[\beta_1,\beta_2]}$). 
   }
\label{fig:sampling}
\vspace{-0.5cm}
\end{figure}
\endgroup
Because contrastive learning \cite{tripletlossfacenet} is highly sensitive to sample mining, we carefully adjust the gap between positive and negative samples to prevent the selection of easy negatives throughout all training phases. We begin by extracting negative samples that are far from the positive ones and gradually tighten the classification difficulty by reducing the distance gap between positives and negatives. To sample positive and negative feature sets, we use the reference features from the left tile $\mathcal{X}$ and a disparity ground truth map $\mathcal{D}$ as follows (see \cref{fig:sampling}):
\begingroup
\setlength{\abovedisplayskip}{3pt}
\setlength{\belowdisplayskip}{3pt}
\begin{equation}
\left\{
\begin{array}{lll}
\mathcal{X}_{+} = \mathcal{X} \pm  ( \mathcal{D} + \mathcal{U_{\interval[-\alpha,\alpha]}} ) \\ \mathcal{X}_{-} = \mathcal{X} \pm  ( \mathcal{D} + \mathcal{U}_{\interval[\beta_1,\beta_2]} ) \\ \alpha < \beta_{1} < \beta_{2}
\end{array}
\right\},
\end{equation}
\endgroup
where $\mathcal{U_{\interval[-\alpha,\alpha]}}$ and $\mathcal{U}_{\interval[\beta_1,\beta_2]}$ are uniform distributions of matching and non-matching sampled positions, respectively; $\alpha$ denotes a symmetric interval around the ground truth for positive samples whereas $\beta_1$ and $\beta_2$ are the negative sampling interval bounds.
\vspace{-0.7cm}
\paragraph{Occlusion handling. }
As our main task is to learn dense similarities by means of binary classification, occlusions should be handled carefully since the correspondence problem is violated in these regions. In patch-based learning approaches, it is addressed by training exclusively on samples extracted in non-occluded areas {\cite{zbontar2016,visualembedding,Han2015}}, {while traditional correlation-based dense matching yields occlusion masks by applying a hard threshold on the computed correlation values \cite{micmac}. 
This approach is effective if similarities in occlusions remain low, which is true for NCC or census similarity metrics. However, for our deep-learnt similarity, resemblance is more than visual and is deduced from the entire feature map structure. Hence, the model should be explicitly told that some features do not have their corresponding matches.} We account for occlusion by labelling sample features extracted from these regions as negatives. By reformulating our training losses, the network is not incentivized to match features in occluded areas which we accomplish by penalizing the inferred similarity measures during training. Put differently, the network is trained to output low similarity scores in occlusion regions. Note that our aim is to filter out occluded areas through similarity and not to enhance the underlying surface regularity. Our new losses are reformulated as follows:
\begingroup
\setlength{\abovedisplayskip}{2pt}
\setlength{\belowdisplayskip}{2pt}
\begin{equation}
\noindent
\label{eq:triocc}
\begin{split}
    \mathcal{L}_{3_{All}}  & = \mathcal{L}_{3_{nocc}} +  \mathcal{L}_{3_{occ}} \\ & =\hspace{-0.5cm} \sum_{(i,j) \in \mathcal{X}_{nocc}}\hspace{-0.5cm}\mathcal{O}(\mathcal{S}^{i,j}_{-}- \mathcal{S}^{i,j}_{+}+ m ,0)\\ & +\hspace{-0.5cm}\sum_{(i,j) \in \mathcal{X}_{occ}}\hspace{-0.25cm}\mathcal{O}(\mathcal{S}^{i,j}_{1-}+\mathcal{S}^{i,j}_{2-},0)~,
\end{split}
\end{equation}
\endgroup
where $\mathcal{S}^{i,j}_{1-}=\left<\mathcal{X}^{i,j}_{1-},\mathcal{X}^{i,j}_{occ}\right>$ and $\mathcal{S}^{i,j}_{2-}=\left<\mathcal{X}^{i,j}_{2-},\mathcal{X}^{i,j}_{occ}\right>$ are cosine similarities between a reference feature $\mathcal{X}^{i,j}_{occ}$ located at occlusions and features $\mathcal{X}^{i,j}_{1-}$ and $\mathcal{X}^{i,j}_{2-}$ sampled from the right feature map. The BCE loss is expressed accordingly:
\begingroup
\setlength{\abovedisplayskip}{2pt}
\setlength{\belowdisplayskip}{1pt}
\begin{equation}
\label{eq:bceocc}
\begin{split}
    \mathcal{L}_{BCE_{All}}  & = \mathcal{L}_{BCE_{nocc}} +  \mathcal{L}_{BCE_{occ}} \\ & \hspace{-0.5cm} =-\hspace{-0.5cm}\sum_{(i,j) \in \mathcal{X}_{nocc}}\hspace{-0.25cm}\mathcal{Y}^{i,j}_{-} \log(1-\mathcal{S}^{i,j}_{-}) + \mathcal{Y}^{i,j}_{+} \log(\mathcal{S}^{i,j}_{+})\\ & \hspace{-0.5cm} -\hspace{-0.5cm}\sum_{(i,j) \in \mathcal{X}_{occ}}\hspace{-0.25cm}\mathcal{Y}^{i,j}_{-} \Bigl(\log(1-\mathcal{S}^{i,j}_{1-}) + \log(1-\mathcal{S}^{i,j}_{2-})\Bigr)~,
\end{split}
\end{equation}
\endgroup
%
%
\subsection{Learning strategy}
\vspace{-0.15cm}
We apply the same sampling scheme to both the feature backbone and the decision network training, alternating between the triplet and the BCE loss.
We adopt a differential learning strategy to train the whole model. More specifically, we set the initial learning rate to \textit{0.001} and progressively divide it by a factor of 10 for later (decoder), intermediate (bottleneck) and earlier (encoder) backbone parameters. 
Following a coarse to fine training scheme,
we set the matching pixel locations sampling interval $\alpha$ to $\{1,0\}$ and the non-matching pixel locations sampling intervals  $\beta_{1}$ and $\beta_{2}$ progressively to $\{2,8\}$, $\{2,6\}$, $\{1,5\}$ and $\{1,4\}$, respectively. This gradual tightening scheme allows to leverage easy negatives at the beginning of training and helps the network learn fast. Then, we track harder negatives by reducing the distance to the ground truth locations. By doing so, we incite that features or learned similarities are not only distinctive far away from correct matches but also within their vicinity. 

All training scenarios are run for 50 epochs per each sampling interval. Finally, we perform a last full tight configuration training (i.e., backbone, MLP, $\alpha=0$, $\beta_{1}=1$, $\beta_{2}=4$). To avoid overfitting, we train our model on tile subsets that are randomly extracted in the course of training. This guarantees that the model sees a quasi-different sample of the dataset at each epoch. 
Note that all models have been initially trained on non-occluded masked areas. The occlusion sensitive loss functions (\cref{eq:triocc,eq:bceocc}) were engaged in the very final training.
\vspace{-0.2cm}
\section{Experiments}\label{sec:experiments}
\vspace{-0.13cm}
\paragraph{Implementation details.}
\begin{figure*}
\centering
   \includegraphics[width=0.8\textwidth]{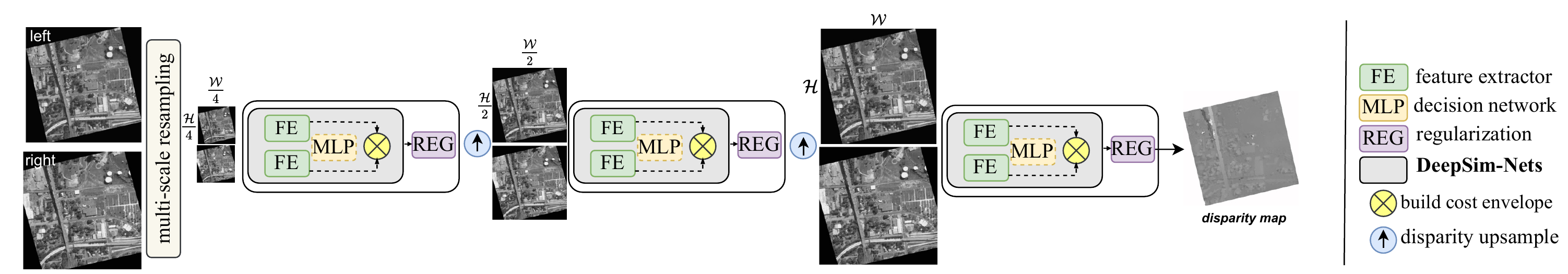}
   \vspace{-0.27cm}
   \caption{\textbf{Multi-Resolution Inference.} We display a $\textit{3-scale}$ iteration. Up to image resolutions $\frac{\mathcal{H}}{8}\times\frac{\mathcal{W}}{8}$, NCC estimates coarse yet robust disparity maps. Moving from resolutions $\frac{\mathcal{H}}{4}\times\frac{\mathcal{W}}{4}$, the down-scaled stereo pairs are iteratively fed to the feature extractor taking one of the \underline{three} variants: \Unet32,\Unet Attention, \textbf{MS-AFF}. We fill the flexible per-pixel disparity range cost structure either with raw cosine of angles between embeddings ($\large\dashrightarrow$) or with the learnt MLP-based similarities. Then, we upscale the predicted disparity map serving as a predictor for the next iteration.}
   \label{fig:epipolarinference}
\vspace{-0.65cm}
\end{figure*}
DeepSim-Nets predict similarities which are used as input to a semi-global matching in the post-processing. The goal is to reduce the underlying noise and penalize disparity jumps within a local neighborhood of the cost structure. This regularization is performed using MicMac's SGM implementation~\cite{micmac}. 
To keep inference memory-friendly and fit for large scale production pipelines, our architecture is integrated into a multi-resolution iterative approach, also present in~\cite{micmac} (see \cref{fig:epipolarinference}). 
%
This approach involves exploring \textit{n-scale} images drawn from the original full resolution image and generating multi-scale aggregated features through concatenation or self-attention mechanism (see \cref{sec:representation}). Our learning models are activated from scale 3. More explicitly, given a full resolution epipolar pair of dimensions $\mathcal{H}\times\mathcal{W}$, our models are deployed from resolution $\frac{\mathcal{H}}{4}\times\frac{\mathcal{W}}{4}$. 
We currently train DeepSim-Nets on a mix of 8-bit and 16-bit single-channel images because our focus is 3D reconstruction and high-resolution satellite sensors are by design panchromatic. However, the network can easily be extended to more channels.\\
\begingroup
\setlength{\intextsep}{0pt}%
\setlength{\columnsep}{0pt}%
\begin{figure}[ht!]
\hspace{-0.2cm}
\centering
\begin{minipage}{0.25\textwidth}
\centering
    \hspace{-1.0cm}
    \centerline{\includegraphics[width=0.9\columnwidth]{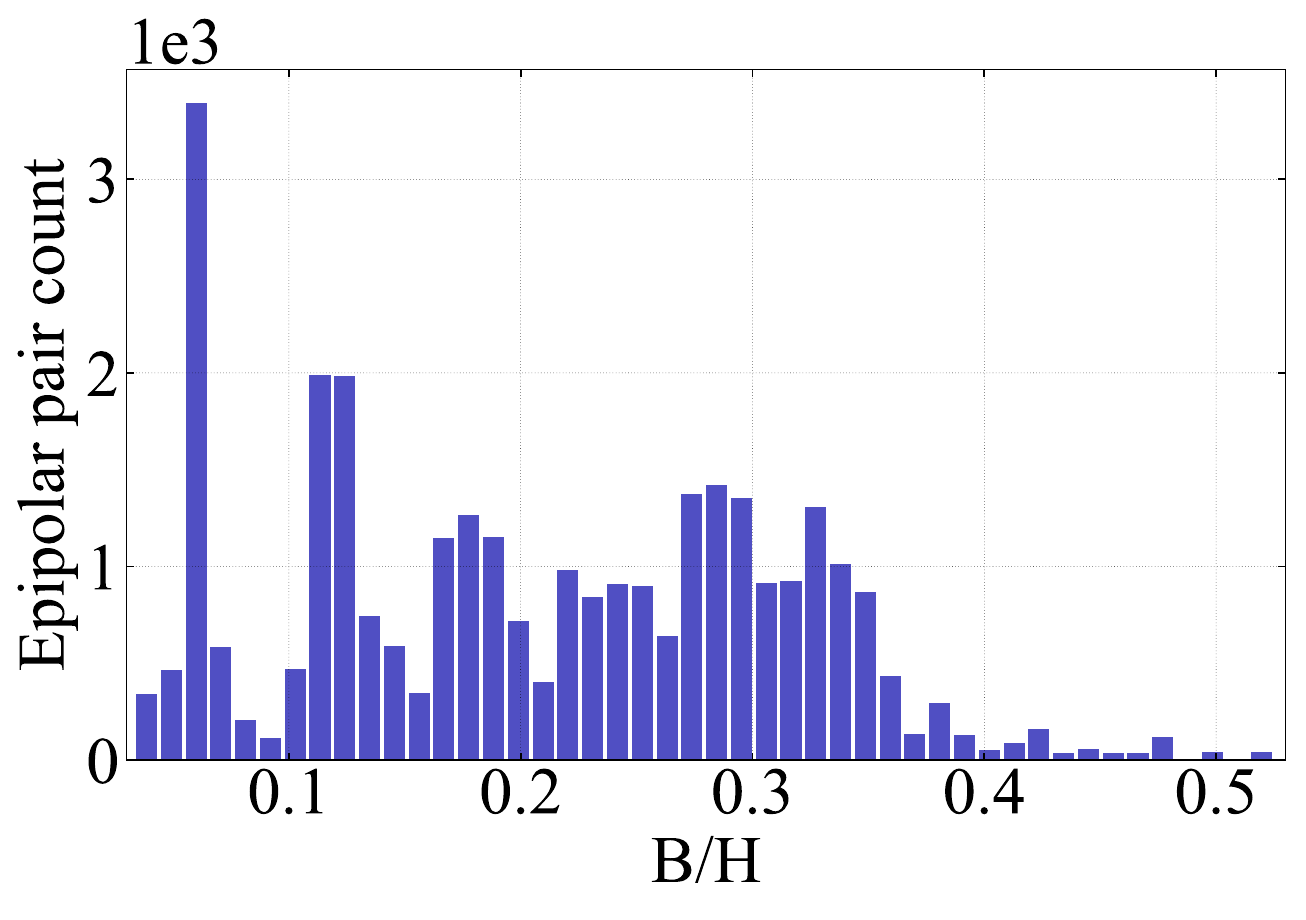}}
    \vspace{-10pt}
    \caption{Training dataset base-to-height ratio distribution.}
    \label{fig:bhtraining}
\end{minipage}
\hspace{0.2cm}
\begin{minipage}{0.18\textwidth}
\vspace{-0.1cm}
\centering
\captionsetup{type=table} 
\scriptsize
\caption{Models parameters setting.}
\vspace{-0.25cm}
\scalebox{0.9}{
  \begin{tabular}{@{}l@{}c@{}}
    \toprule
    Model & $\#$ of params. ($\times10^3    $) \\
    \midrule
    MC-CNN\cite{zbontar2016} & 148\\
    MS-AFF(\textbf{Ours}) & 965\\
    \Unet32 & 7,800\\
    \Unet Attention & 9,500\\
    MLP & 345\\
    \bottomrule
  \end{tabular}
  }
  \label{tab:modelparams}
\end{minipage}
\vspace{-0.65cm}
\end{figure}
%
\hspace{-0.1cm}\noindent\textbf{Datasets.}\label{datasets} We perform training on the aerial dataset \cite{dublincity,wuteng} consisting of \textit{30,841}; \textit{3164} and \textit{607}  pairs of epipolar images over Dublin, Enschede and Vaihingen, respectively. All tiles sizes are set to $\textit{1024} \times \textit{1024}$. A ground truth reprojected \Lidar disparity map is given for each epipolar pair. Evaluation is performed on aerial stereo pairs (Ground Sampling Distance  GSD=8cm) over Toulouse, satellite stereo pairs over Buenos Aires (WV-3, GSD=30 cm) and Montpellier (Pléiades 1B, GSD=50 cm). {The Toulouse dataset is closely related to our training dataset, as they share similar sensor characteristics and spatial resolutions.
On the other hand, the satellite stereo pairs with their specific acquisition geometry, spatial resolution and low signal-to-noise ratio can be considered as \textit{out-of-distribution} datasets. The satellite datasets are therefore appropriate for benchmarking the transferability of our method.
}
\endgroup\\
\noindent\textbf{Metrics.} \label{sec:metrics}Evaluation metrics for binary classifiers performance assessment include accuracy, confusion matrix as well as ROC curves, capturing recall at different decision boundaries. Since we target sufficiently separable matching and non-matching feature populations for the subsequent regularization task, we do not privilege a certain threshold. Instead, we estimate the joint probability distribution by sampling matching and non-matching pixel locations at different interval settings (see \cref{fig:jp}). A perfect classifier yields matching similarities that are always greater that non-matching ones. We compute joint probability area under the diagonal, denoted as \textit{JP} and the marginal distributions geometric intersection area denoted as \textit{InterA} (see \cref{tab:statsclassif}). We also provide \textit{n-pixel} error histograms, 1-, 2- and 3-pixel errors and compare our DeepSim-Nets with MC-CNN~acrt and PSMNet in \cref{fig:disparityhistograms} and \cref{tab:errorrates}. Note that PSMNet was trained on a larger aerial dataset acquired on various cities~\cite{wuteng}, including those used for training our models.\par
\vspace{0.2cm}
\noindent\textbf{Ablations.} Three modelling hypothesis are validated: (1)~\textit{no$\_$occlusion} loss term contribution, (2) transferability of our model trained on aerial images to a satellite configuration, and (3) the contribution of the MLP-learnt similarity compared to the baseline cosine feature-level similarity.
\section{Results and discussion}\label{sec:results}
\begin{figure}[t!]
  \centering
   \hspace{-0.5cm}
   \includegraphics[width=0.5\textwidth]{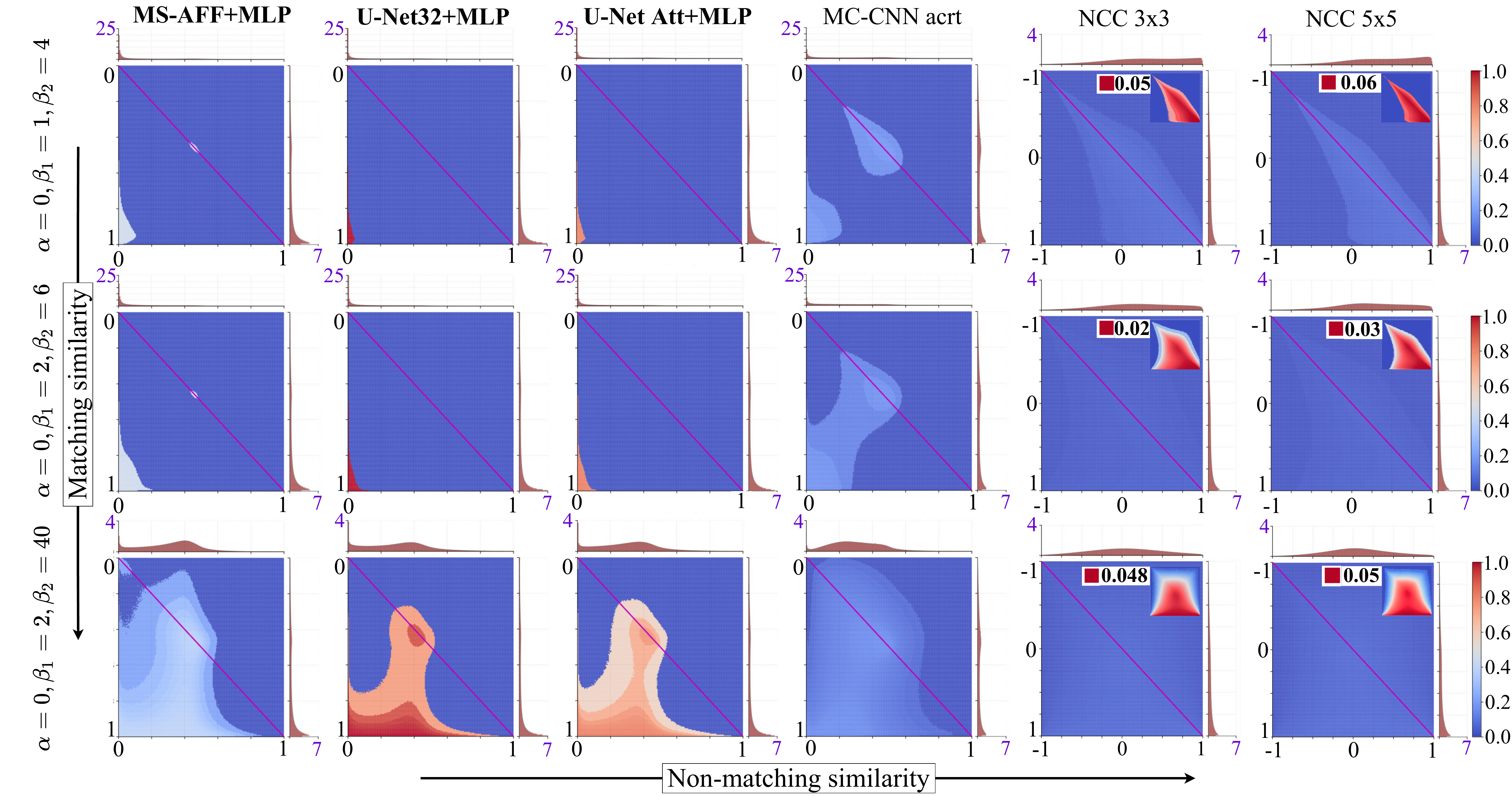}
   \vspace{-0.6cm}
    \caption{\textbf{Classifiers Accuracy}. From top to bottom, we enlarge the negatives' sampling interval defined by $\beta_{1}$ and $\beta_{2}$. We estimate joint as well as marginal matching/non-matching similarity distributions for the \textbf{three} variants of DeepSim-Nets, MC-CNN acrt and NCC with $3\times3$ and $5\times5$ window sizes. For visualization purposes, we normalize all joint distributions \textit{w.r.t} the maximum distribution value in each row and display equalized thumbnails for NCC distributions. These maps give us insights on the matching and non-matching "pixels" separability of our binary classifiers. Our DeepSim-Nets (first 3 columns) accumulate almost all observations under the diagonal. MC-CNN acrt and NCC misclassify and render high variance maps.}
  \label{fig:jp}
  \vspace{-0.2cm}
\end{figure}

\paragraph{Similarity learning.}
Joint probability maps computed on unseen aerial data are visualised in \cref{fig:jp}. The most compact distributions are produced by \Unet32 and \Unet Attention, which are condensed near 1 for matching similarities (abscissa) and 0 for non-matching ones (ordinate). 
The MS-AFF distribution qualitatively follows the same trend, but shows a small blob for similarities equal to 0.5, which indicates that some positives and negatives are hard to classify. However, our MS-AFF model yields the highest \textit{JP} for all sampling scenarios (\cref{tab:statsclassif}), indicating that the misclassified samples population is negligible. When mixing near and far negative samples (see \cref{fig:jp} (row 3) $\&$ \cref{tab:statsclassif} (col. 3)), the variance of our proposed models' joint distributions increases but still outperforms local models by at least 4 $\%$.\par
Local neighborhood models, including NCC $3\times3$, NCC $5\times5$, and MCC-CNN acrt, exhibit an increase in \textit{JP}  and a decrease in \textit{InterA} as the negative sampling intervals increase, whereas our global ensemblistic models follow the opposite tendency. This occurs because window-based methods enhance the feature's distinctiveness by moving further away from correct matches, where local neighborhood changes drastically and pixel classification becomes easier. On the other hand, our DeepSim-Nets are designed to be distinctive near correct matches, while on large unexplored negative sampling intervals, they may misclassify.
The ROC curves in \cref{fig:auc} reveal that  our models yield higher recall rates compared to MC-CNN acrt. MS-AFF performs marginally worse than \Unet32 and \Unet Attention as the AUC demonstrates in \cref{tab:statsclassif}. With the model complexity kept in mind, our lightweight MS-AFF shows decent classification results across all examined metrics.
\begingroup
\setlength{\intextsep}{0pt}
\setlength{\tabcolsep}{2.5pt}
\begin{table}[t!]
  \centering
  \scriptsize
  \caption{\textbf{Quantitative Evaluation.} We evaluate Models' similarity classification performance on unseen aerial Toulouse dataset. Our DeepSim-Nets outperform local methods: MC-CNN acrt, NCC. Our lightweight \textbf{MS-AFF} yields the highest \textit{JP} for all sampling scenarios and performs marginally worse than \Unet32 and \Unet Attention on AUC and \textit{InterA}. }
  \vspace{-0.3cm}
  \scalebox{0.92}{
  \begin{tabular}{@{}lccc@{}|ccc@{}|ccc@{}}
    \toprule
    Sample setting & \multicolumn{3}{c|}{$\beta_{1}=1,\beta_{2}=4$} & \multicolumn{3}{c}{$\beta_{1}=2,\beta_{2}=6$} & \multicolumn{3}{|c}{$\beta_{1}=2,\beta_{2}=40$}\\
    \midrule
    \multicolumn{9}{c}{Aerial Test dataset, $\alpha=0$}\\
    \midrule
    Metrics ($\%$) & JP$\uparrow$   &  InterA$\downarrow$  & AUC$\uparrow$  &  JP$\uparrow$   &  InterA$\downarrow$  & AUC$\uparrow$  & JP$\uparrow$   &  InterA$\downarrow$  & AUC$\uparrow$ \\
    \textbf{Ours:}& & & & & & & & &\\
    \hspace{0.1cm}\Unet32 + MLP   & 85.3  &  23.6  & 92.2  &  88.6  & 15.9 & 95.5  & 87.3 & 19.1 & 89.8 \\
    \hspace{0.1cm}\Unet Att. + MLP & 85.4  &  \textbf{23.3}  & \textbf{92.2} &  88.9  & \textbf{15.4} & \textbf{95.7}  & 88.0 & 18.7 & \textbf{90.1} \\
    \hspace{0.1cm}MS-AFF + MLP & \textbf{86.4}  &  23.6  & 91.4  &  \textbf{89.6}  & 15.7 & 95.2 & \textbf{88.0} & \textbf{18.4} & 89.6\\
    \cline{1-10}
    MC-CNN acrt \cite{zbontar2016} & 78.0  &  33.6  & 85.3 & 82.0   & 25.3 & 89.8  & 83.7 & 24.9 & 87.3 \\
    NCC (3$\times$3) & 71.2  &  76.7  & -- & 74.8 & 59.1 & --  & 77.0 & 45.4 & -- \\
    NCC (5$\times$5) &  73.2  &  75.4  & -- & 76.1 & 60.3 & --  & 80.0 & 40.6 & -- \\
    \bottomrule
  \end{tabular}}
  \label{tab:statsclassif}
  \vspace{-0.5cm}
\end{table}
\endgroup
DeepSim-Nets overcome local methods matching ambiguities near correct matches and leverage decent distinctive similarities that are mandatory for the subsequent surface reconstruction. Moreover, although not trained on large negatives sampling offsets, salient {matching} similarities are obtained when $\beta_{1}=2,\beta_{2}=40$. We also achieve pixel-level separability as well as matching coherence for homogeneous areas.

By explicitly labeling correspondences computed over occlusions as negative samples, we encourage dissimilarity across these regions. This in turn facilitates occlusion detection through simple thresholding of the similarity map. \cref{fig:occ_non_occ} illustrates this behaviour with MS-AFF trained without and with occlusion self-supervision.
\begingroup
\setlength{\intextsep}{0pt}%
\begin{figure}[t!]
\hspace{-0.2cm}
  \centering
   \includegraphics[width=1.0\linewidth]{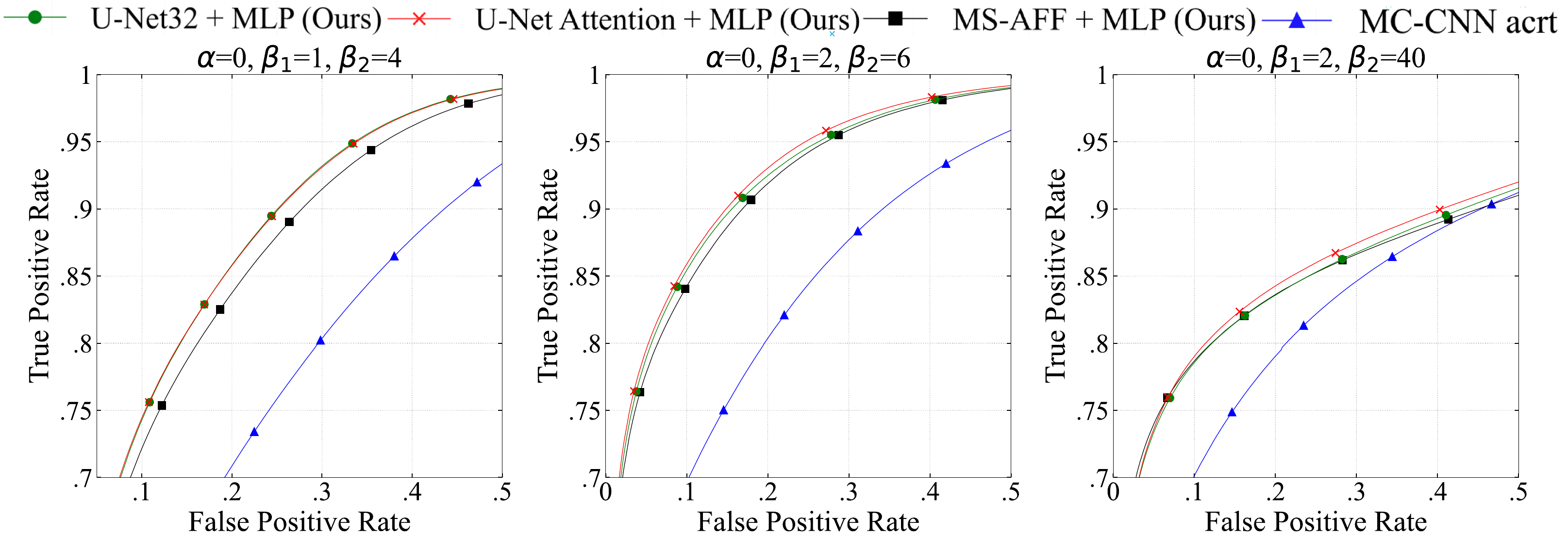}
   \vspace{-0.2cm}
    \caption{\textbf{ROC Curve Analysis}. We represent ROC curves for \emph{three} negatives sampling intervals (i.e scenarios) defined by offsets $\beta_{1}$ and $\beta_{2}$ \textit{w.r.t} ground truth locations ($\alpha=0$). Our ensemble models yield the lowest False Positive Rates (FPR) for different recall rates compared to MC-CNN acrt for \underline{all} sampling scenarios.}
  \label{fig:auc}
  \vspace{-0.2cm}
\end{figure}
\endgroup
\begingroup
\setlength{\intextsep}{0pt}%
\begin{figure}[h!]
  \centering
   \includegraphics[width=0.85\linewidth]{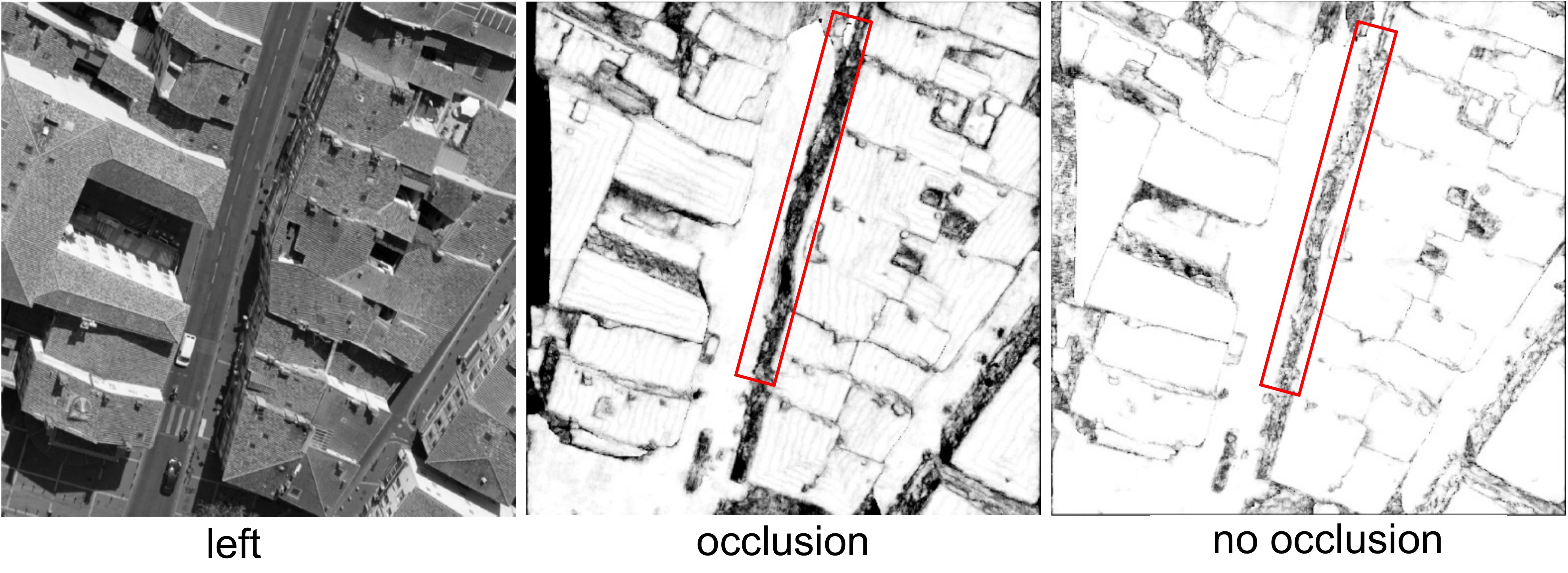}
   \vspace{-0.35cm}
    \caption{\textbf{Contribution of The Occlusion Term}. When the occlusion-specific loss term is activated, similarity values drop drastically in occluded regions \textcolor{red}{$\square$}.}
  \label{fig:occ_non_occ}
\vspace{-0.7cm}
\end{figure}
\endgroup
\begin{table}[t!]
 \centering
   \scriptsize
  \caption{\textbf{Ablation Study. }Accuracy assessment on aerial/satellite {unseen} datasets. Statistics are calculated on difference maps between models' induced disparities and ground truth in \underline{non-occluded} areas. For the aerial Toulouse dataset, we examine varying $\frac{B}{H}$. For Montpellier, Pléiades 1B dataset, no $\frac{B}{H}$ selection is done. $\mu$ is the mean absolute difference, $\sigma$ is the standard deviation, $\textit{NMAD}$ is the normalized median absolute deviation. $D_{1}$, $D_{2}$ and $D_{3}$ are 1-, 2- and 3-pixel error rates ($\%$) respectively.}
  \vspace{-0.1cm}
  \begin{tabular}{@{}l@{}cccccccc@{}}
    \toprule
    Method & $\frac{B}{H}$  & $\mu$$\downarrow$ & $\sigma$$\downarrow$ & $\textit{NMAD}$$\downarrow$ & $D_{1}$$\downarrow$ & $D_{2}$$\downarrow$ & $D_{3}$$\downarrow$ \\
    \midrule
    \multicolumn{8}{c}{Toulouse aerial 8cm GSD dataset}\\
    \midrule
    \hspace{0.05cm}$|$\hspace{0.13cm} MS-AFF \text{cos} & \multirow{7}{*}{0.2}  & 0.42 & 0.98 & 0.12 & 7.10 & 3.43 & 2.40\\
    \begin{turn}{90}\textbf{S}\end{turn}\hspace{0.1cm} \Unet32 \text{cos}&   & 0.40 & \textbf{0.96} & 0.11 & 6.61 & 3.26 & 2.30\\
    \begin{turn}{90}\textbf{R}\end{turn}\hspace{0.1cm} \Unet-Attention \text{cos}&   & \textbf{0.40} & 0.96 & \textbf{0.11} & \textbf{6.40} & \textbf{3.19} & \textbf{2.28}\\
    \cline{3-8}
    \begin{turn}{90}\textbf{U}\end{turn}\hspace{0.1cm} MS-AFF+MLP &  & 0.39 & 0.95 &  \textbf{0.11} & 6.30 & 3.22 & 2.28\\
    \begin{turn}{90}\textbf{O}\end{turn}\hspace{0.1cm} \Unet32+MLP  &  &  0.39 & \textbf{0.95} & 0.11 & 6.18 & 3.13 & 2.24\\
    \hspace{0.05cm}$|$\hspace{0.13cm} \Unet-Attention+MLP & & \textbf{0.39} & 0.95 & 0.11 & \textbf{5.97} & \textbf{3.10} & \textbf{2.24} \\
    \cline{3-8}
    MC-CNN acrt\cite{zbontar2016} & & 0.52 & 1.14 & 0.14 & 9.88 & 5.23 & 3.60\\
    PSMNet\cite{PSMNet} & & 0.49 & 1.00 & 0.15 & 8.66 & 4.60 & 3.01 \\
    \toprule
    \hspace{0.05cm}$|$\hspace{0.13cm}  MS-AFF \text{cos} & \multirow{7}{*}{0.48}  & 1.18 & 1.61 & \textbf{0.37} & \textbf{31.37} & 13.62 & 9.09\\
    \begin{turn}{90}\textbf{S}\end{turn}\hspace{0.1cm} \Unet32 \text{cos}&   & 1.19 & 1.58 & 0.38 & 32.66 & 13.38 & 8.67\\
    \begin{turn}{90}\textbf{R}\end{turn}\hspace{0.1cm} \Unet-Attention \text{cos}&   & \textbf{1.18} & \textbf{1.58} & 0.38 & 32.34 & \textbf{13.23} & \textbf{8.60}\\
    \cline{3-8}
    \begin{turn}{90}\textbf{U}\end{turn}\hspace{0.1cm} MS-AFF+MLP &  & 1.27  & \textbf{1.49} &  \textbf{0.40} & \textbf{39.24} & \textbf{14.60} & 8.65\\
    \begin{turn}{90}\textbf{O}\end{turn}\hspace{0.1cm} \Unet32+MLP  &  &  1.28 & 1.50 & 0.42 & 40.35 & 14.75 &  8.44\\
    \hspace{0.05cm}$|$\hspace{0.13cm}  \Unet-Attention+MLP & & \textbf{1.27} & 1.50 & 0.42 & 39.56 & 14.77 & \textbf{8.41}\\
    \cline{3-8}
    MC-CNN acrt\cite{zbontar2016} & & 2.10 & 1.96 & 0.99 & 60.31 & 39.73 & 23.45\\
    PSMNet\cite{PSMNet} & & 1.34 & 1.65 & 0.44 & 39.10 & 16.68 & 10.42 \\
    \toprule
    \hspace{0.05cm}$|$\hspace{0.13cm}  MS-AFF \text{cos} & \multirow{7}{*}{All} & 0.70  & 1.30 & 0.21 & 15.97 & 7.15 & 4.85\\
    \begin{turn}{90}\textbf{S}\end{turn}\hspace{0.1cm} \Unet32 \text{cos}&   & 0.69 & 1.28 & 0.21 & 16,15 & 6.97 & 4.63\\
    \begin{turn}{90}\textbf{R}\end{turn}\hspace{0.1cm}  \Unet-Attention \text{cos}&   & \textbf{0.68} & \textbf{1.28} & \textbf{0.20} & \textbf{15,88} & \textbf{6.86} & \textbf{4.59}\\
    \cline{3-8}
    \begin{turn}{90}\textbf{U}\end{turn}\hspace{0.1cm}  MS-AFF+MLP &  & 0.71  & \textbf{1.25} & 0.22 & 18.20 & 7.33 & 4.57\\
    \begin{turn}{90}\textbf{O}\end{turn}\hspace{0.1cm}  \Unet32+MLP &  &  0.72 & 1.26 & 0.22 & 18.59 & 7.35 &  4.50\\
    \hspace{0.05cm}$|$\hspace{0.13cm}  \Unet-Attention+MLP & & \textbf{0.71} & 1.25 & \textbf{0.22} & \textbf{18.14} & \textbf{7.32} & \textbf{4.47}\\
    \cline{3-8}
    MC-CNN acrt\cite{zbontar2016} & & 1.09 & 1.66 & 0.29 & 29.12 & 17.65 & 10.74\\
    PSMNet\cite{PSMNet} & & 0.79 & 1.33 & 0.24 & 19.18 & 8.78 & 5.57\\
    \midrule
    \multicolumn{8}{c}{Montpellier Pléaides 1B 50 cm GSD satellite dataset}\\
    \midrule
    \hspace{0.05cm}$|$\hspace{0.13cm}  MS-AFF \text{cos} & \multirow{7}{*}{--} & 0.75  & 0.98 & \textbf{0.27} & 20.07 & 7.9 & 4.20\\  
    \begin{turn}{90}\textbf{S}\end{turn}\hspace{0.1cm} \Unet32 \text{cos}&   & 0.75 & 1.00 & 0.27 & \textbf{19.94} & 7.86 & 4.18\\
    \begin{turn}{90}\textbf{R}\end{turn}\hspace{0.1cm} \Unet-Attention \text{cos}&   & \textbf{0.75} & \textbf{0.97} & 0.28 & 20.27 & \textbf{7.70} & \textbf{3.98}\\
    \cline{3-8}
    \begin{turn}{90}\textbf{U}\end{turn}\hspace{0.1cm} MS-AFF+MLP &  & 1.27  & 1.26 & 0.50 & 45.31 & 17.56 & 8.72\\
    \begin{turn}{90}\textbf{O}\end{turn}\hspace{0.1cm} \Unet32+MLP &  &  1.17 & 1.22 & 0.46 & 39.84 & 15.82 & 7.81\\
    \hspace{0.05cm}$|$\hspace{0.13cm}  \Unet-Attention+MLP & & 1.20 & 1.26 & 0.49 & 40.87 & 17.13 & 8.56\\
    \cline{3-8}
    MC-CNN acrt\cite{zbontar2016} & & 0.84 & 1.13 & 0.30 & 23.13 & 10.14 & 5.54\\
    PSMNet\cite{PSMNet} & & 1.02 & 1.18 & 0.40 & 32.27 & 13.78 & 7.00\\
    \bottomrule
  \end{tabular}
  \label{tab:errorrates}
\end{table}
\begingroup
\vspace{-0.5cm}
\setlength{\intextsep}{0pt}%
\begin{figure}[h!]
\hspace{-0.15cm}
  \centering
   \includegraphics[width=\linewidth]{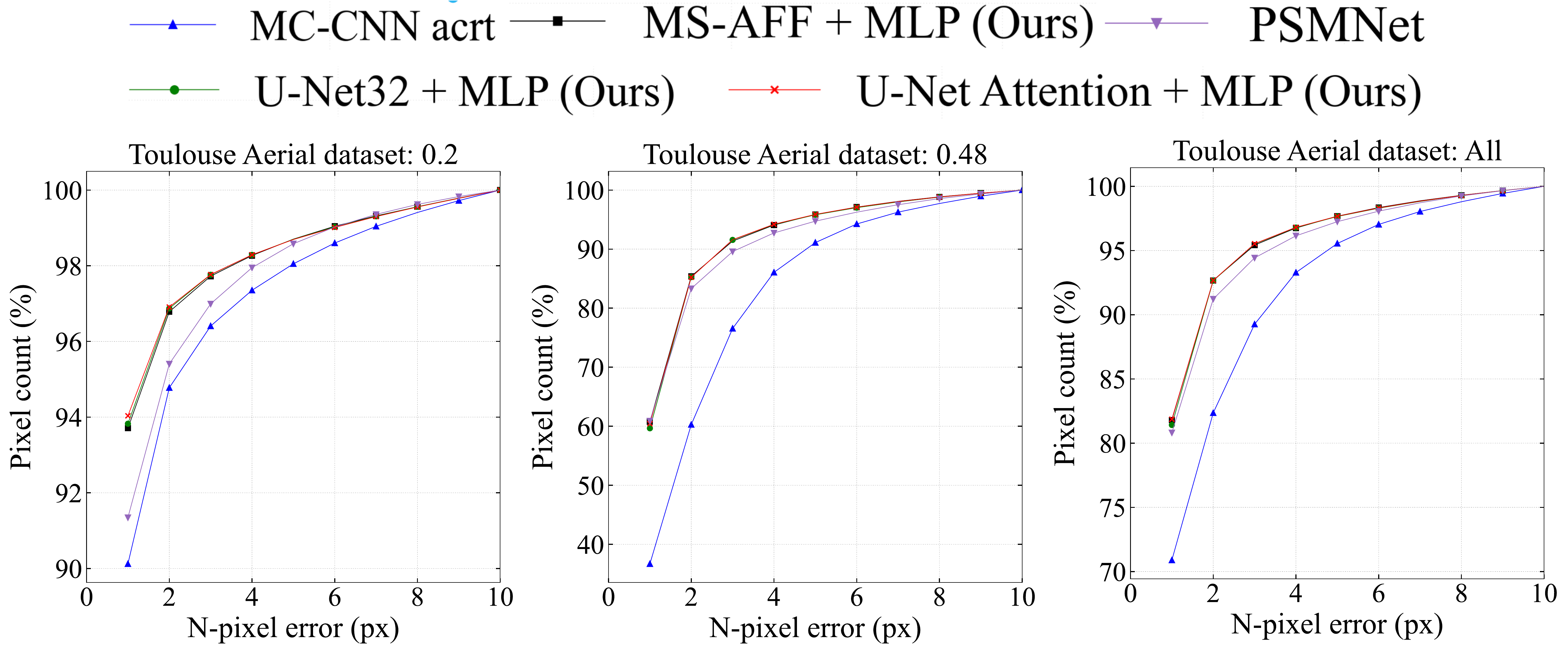}
   \vspace{-0.7cm}
    \caption{\textbf{Quantitative Results}. Error histograms evaluated on \underline{occlusion-free} areas. Our models outperform PSMNet and MC-CNN acrt for different $\frac{B}{H}$ settings.}
  \label{fig:disparityhistograms}
\vspace{-0.6cm}
\end{figure}
\endgroup
\begin{figure*}[tbh!]
  \centering
      \subfloat[Epipolar Pair.]{\includegraphics[height=3.9cm,width=1.95cm]{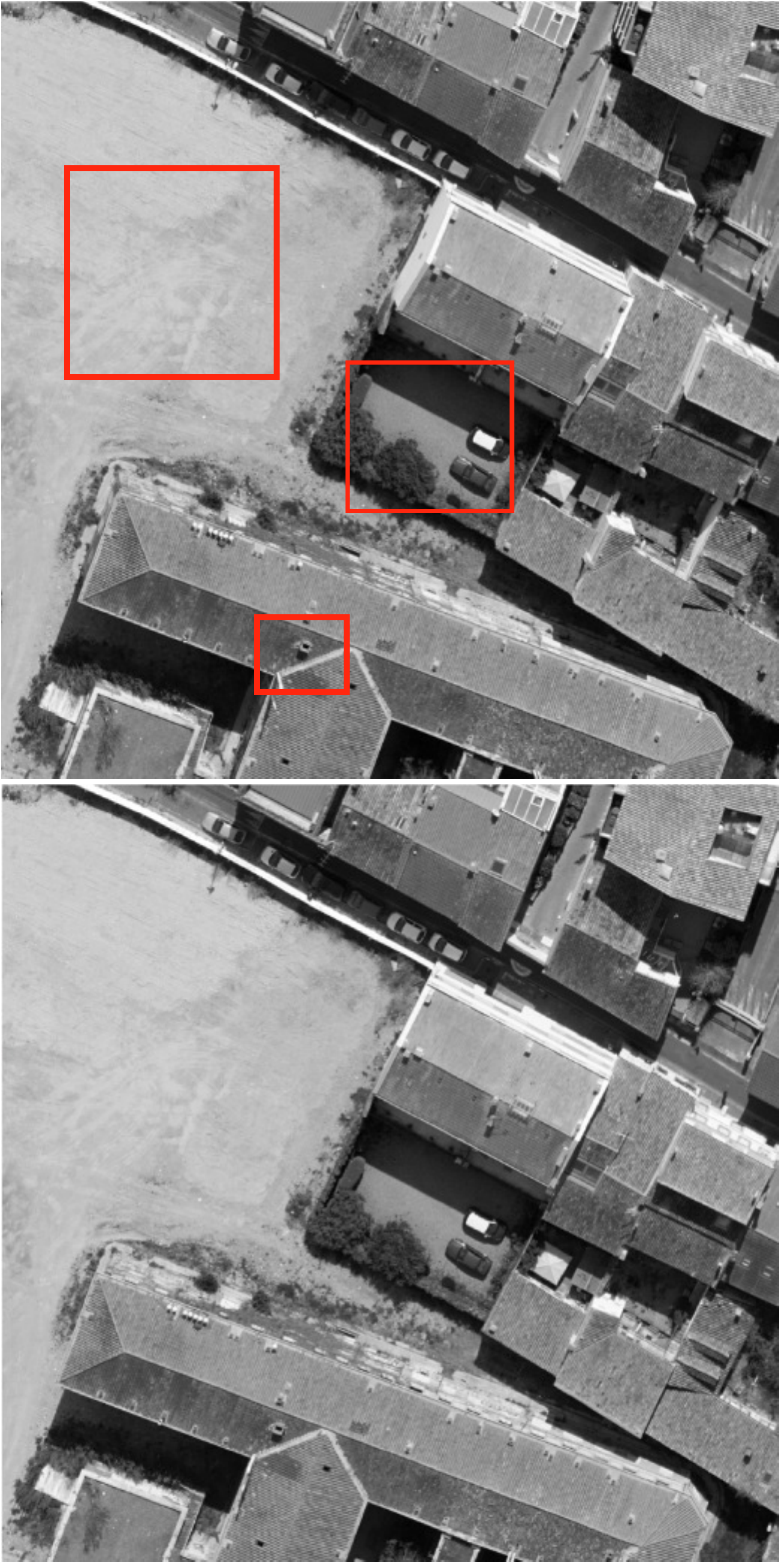}\label{fig:lrepip}\quad}
      \subfloat[\textbf{Ours}:MS-AFF.]{\includegraphics[height=3.9cm,width=1.95cm]{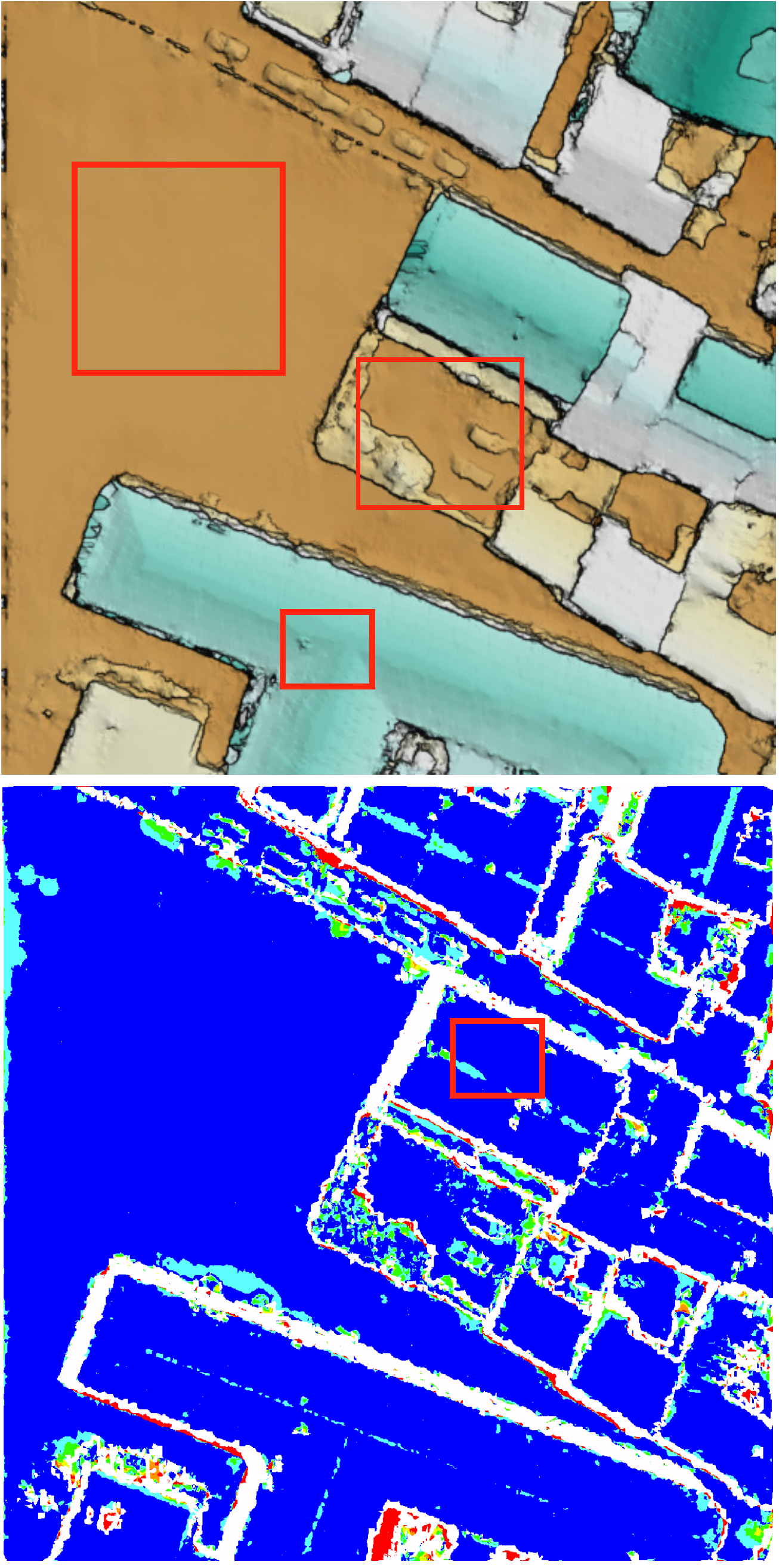}\label{fig:msaffmlp}\quad}
    \subfloat[\textbf{Ours}:\Unet32.]{\includegraphics[height=3.9cm,width=1.95cm]{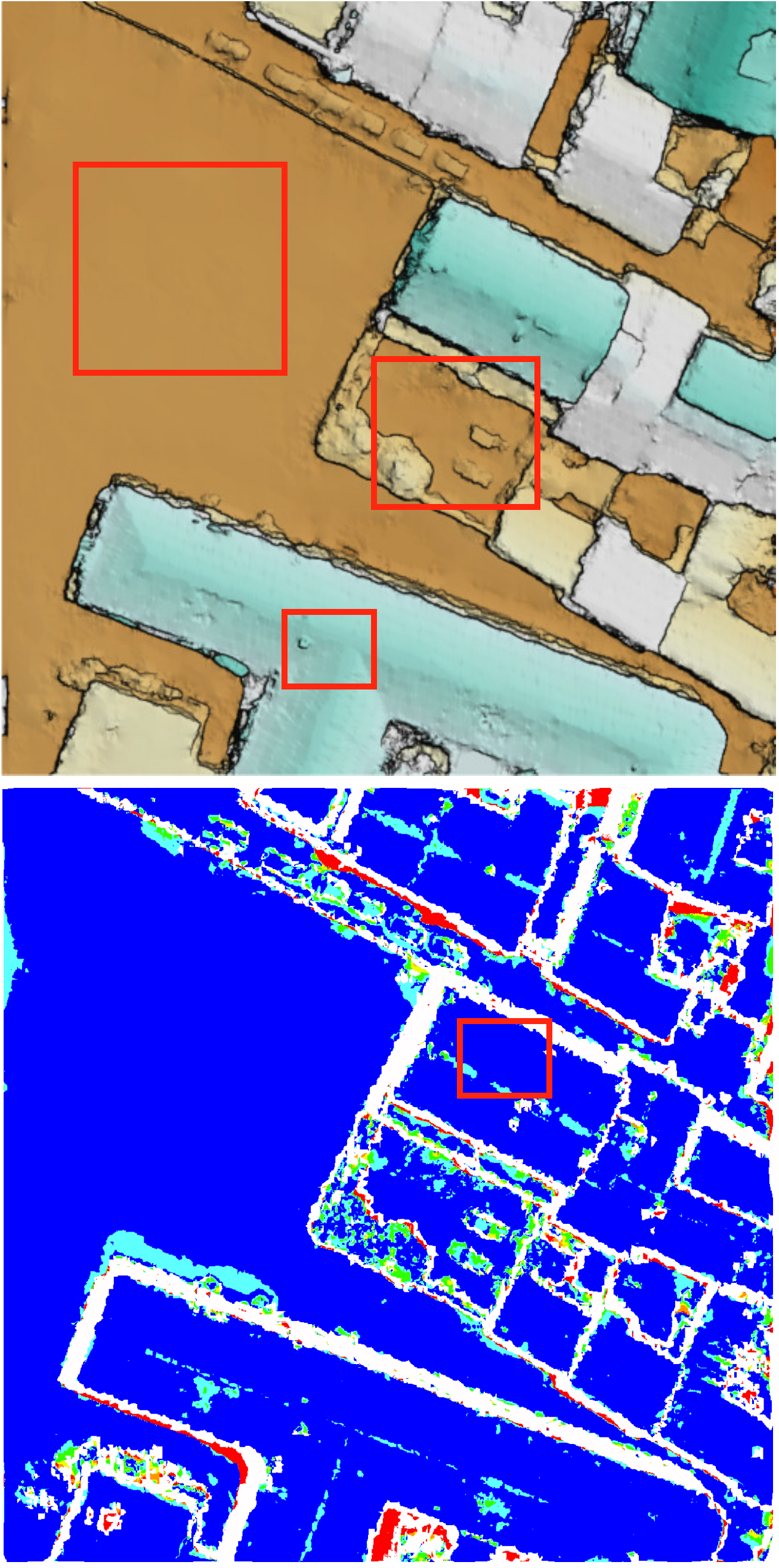}\label{fig:unet32mlp}\quad}
    \subfloat[\textbf{Ours}:\Unet Att.]{\includegraphics[height=3.9cm,width=1.95cm]{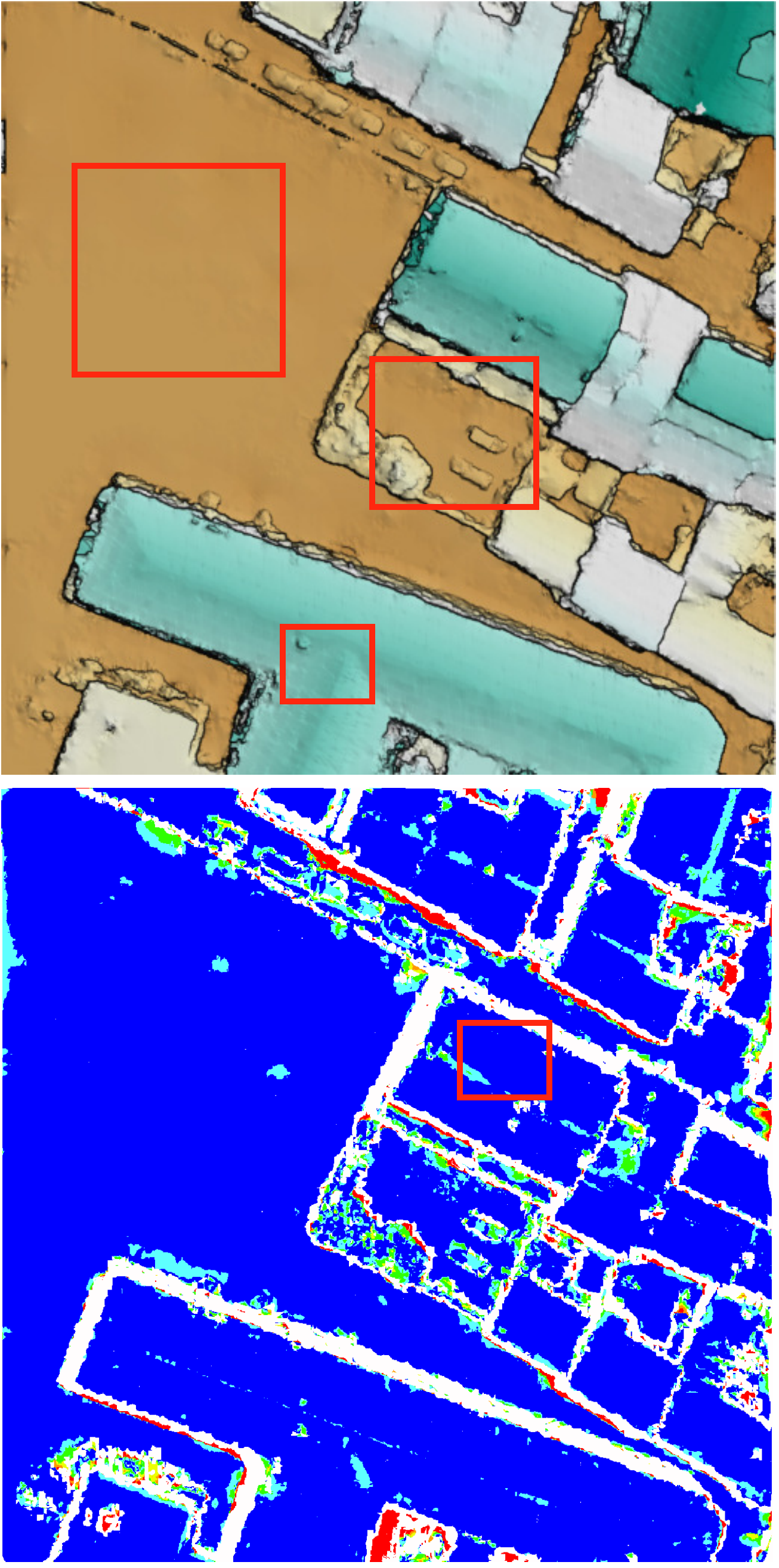}\label{fig:unetattmlp}\quad}
    \subfloat[MC-CNN acrt.]{\includegraphics[height=3.9cm,width=1.95cm]{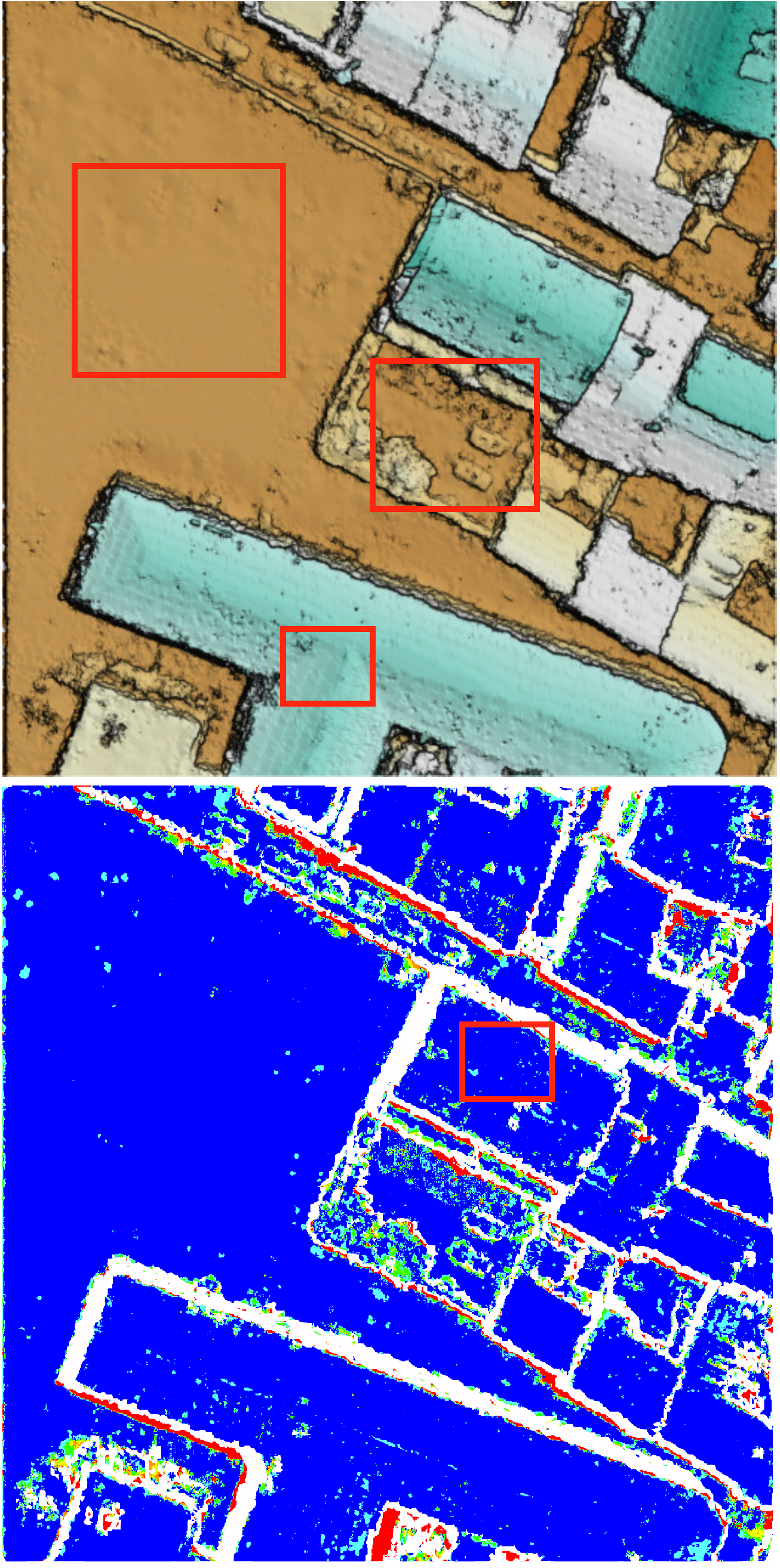}\label{fig:mccnnacrt}\quad}
    \subfloat[NCC$5\times5$.]{\includegraphics[height=3.9cm,width=1.95cm]{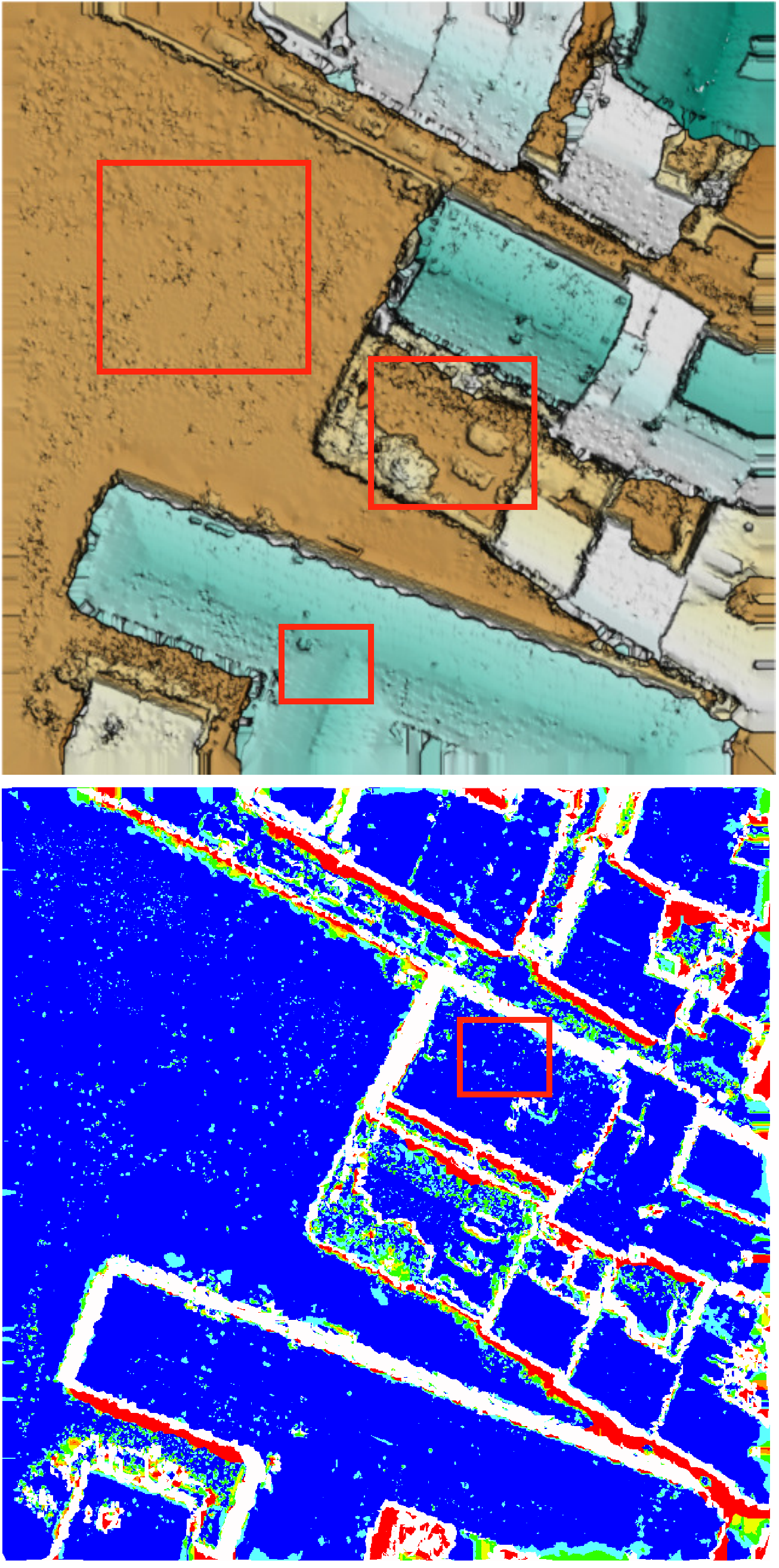}\label{fig:corr}\quad}
    \subfloat[PSMNet.]{\includegraphics[height=3.9cm,width=1.95cm]{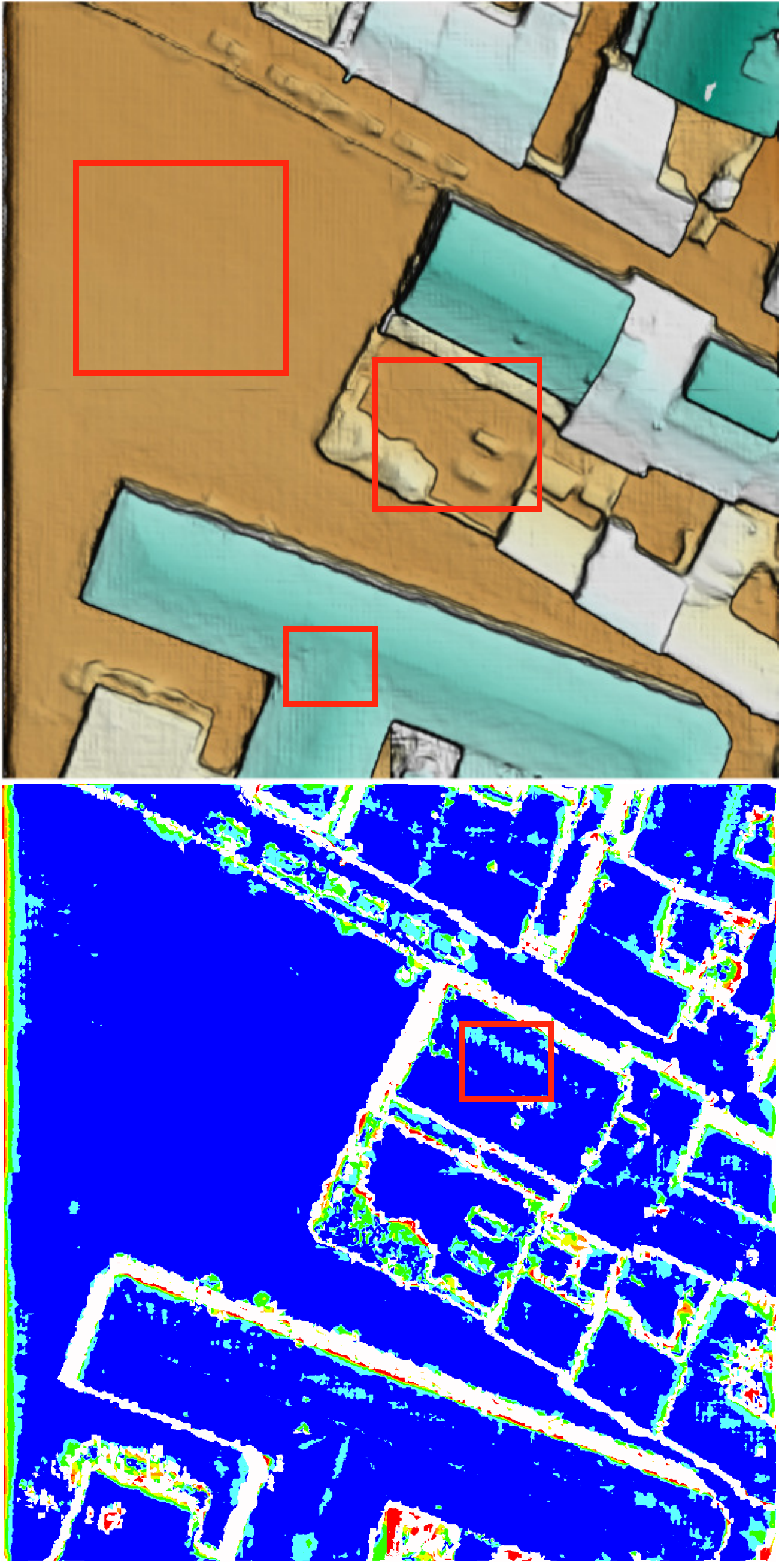}\label{fig:respsmnet}\quad}
    \stackunder[5pt]{\includegraphics[height=2.2cm,width=0.7cm]{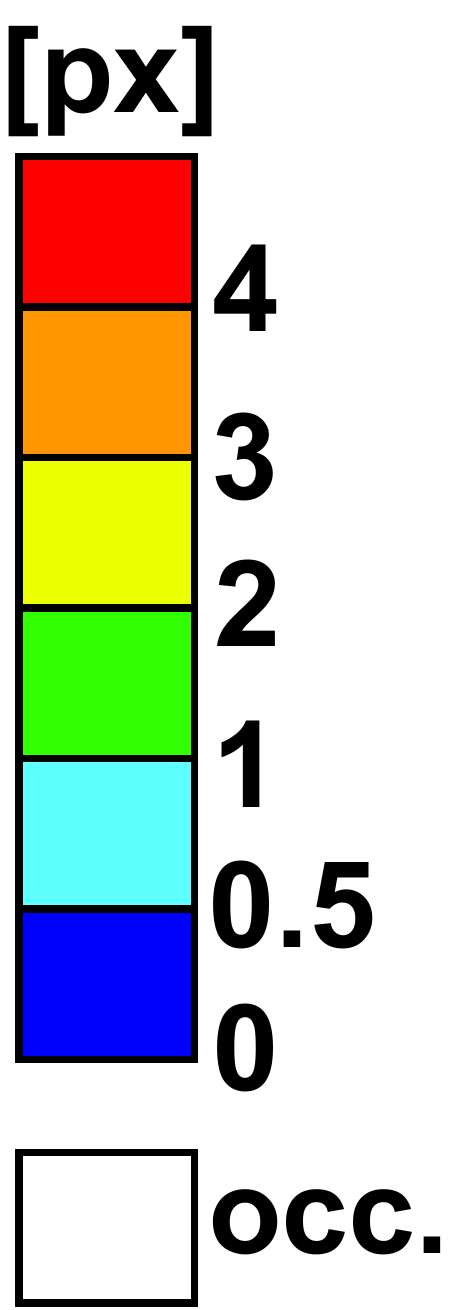}\label{fig:legend}}{}
   \vspace{-0.2cm}
    \caption{\textbf{Disparity Predictions on Aerial Images}. (top: colored and grey-shaded disparity maps, bottom: difference maps \textit{w.r.t} ground truth). Here, we evaluate the entire setting (feature extractor + MLP). Conversely to MC-CNN acrt and NCC, planar surfaces with poor contextual information (big \textcolor{red}{$\square$}) are {recovered best} by our models. {Shadows are handled well by DeepSimNets and PSMNet (middle \textcolor{red}{$\square$}). PSMNet renders consistent reconstructions on buildings boundaries and near occlusions but ignores tiled roof patterns that are recovered by almost all similarity-driven models} (small \textcolor{red}{$\square$}).}  
  \label{fig:disparitytoul}
  \vspace{-0.1cm}
\end{figure*}
\definecolor{mygrrr}{rgb}{0.0,0.6,0.0}
\definecolor{mypurple}{rgb}{1.0,0.0,1.0}
\begin{figure*}[tbh]
\vspace{-0.1cm}
  \centering
    \includegraphics[width=0.95\textwidth]{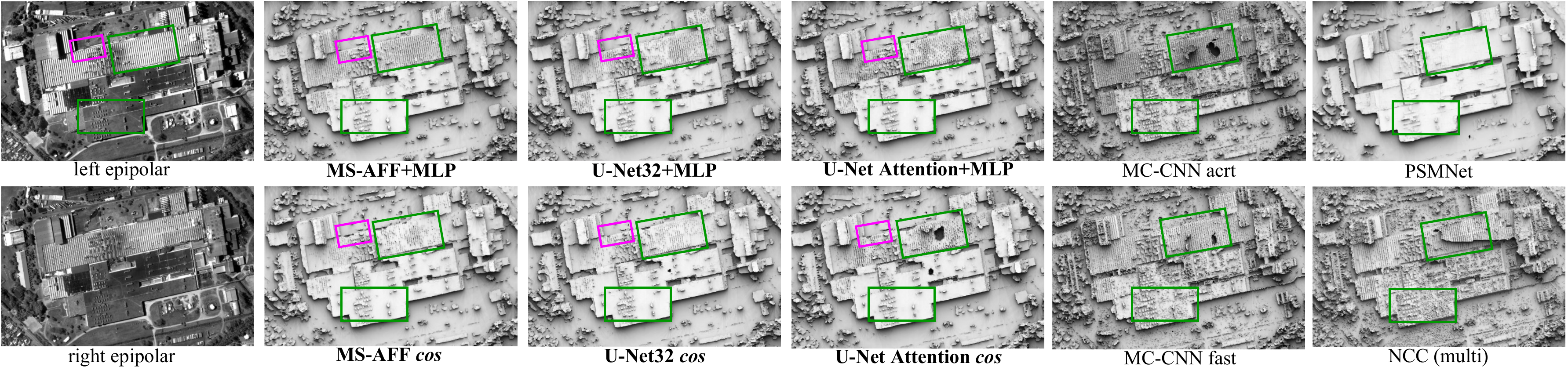}
   \vspace{-0.3cm}
    \caption{\textbf{Disparity Predictions on Satellite Images}. Grey-shaded disparity maps capturing the performance of tested methods on unseen WV3 stereo pairs over Buenos Aires. While local neighborhood classifiers: MC-CNN(fast$\&$acrt),NCC(multiple windows) fail to reconstruct fine-grain building rooftops' details \textcolor{mygrrr}{$\square$}, our models recover such high frequency details. PSMNet acts as a low-pass filter. The MLP-learnt similarities recover buildings' outlines \textcolor{mypurple}{$\square$} missed by raw-cosine (\textit{cos}) similarities.} 
  \label{fig:epipsat_wv3}
  \vspace{-0.5cm}
\end{figure*} 
\paragraph{Surface inference.}
{We compare DeepSim-Nets' disparities against PSMNet and MC-CNN acrt on \emph{unseen} aerial close-to-distribution (Toulouse) and satellite out-of-distribution stereo pairs} (Montpelier, see \cref{tab:errorrates}).  We also study the impact of acquisition geometry on the disparity accuracy by looking at the base-to-height ratio ($\frac{B}{H}$). 
On Toulouse dataset, we show that our models outperform PSMNet and MC-CNN acrt {in occlusion-free regions} almost for all examined metrics (\cref{fig:disparityhistograms} $\&$ \cref{tab:errorrates}). Nonetheless, PSMNet recovers precisely buildings' outlines while our method may render poor edge shapes, especially near occlusions (see \cref{fig:teaser}). This said, PSMNet has the tendency to smooth surfaces and occasionally add high frequency low amplitude artefacts (see \cref{fig:teaser}), while we faithfully reproduce rooftop details (see \cref{fig:disparitytoul}). 
For $\frac{B}{H}=0.2$, \Unet Attention slightly outperforms \Unet32 and MS-AFF on both feature-based \textit{cosine} and MLP-based similarities. The MLP decision module gain is about 0.5 $\%$ for all models. This shows that feature modules provide sufficiently generic representations, deployable for downstream tasks such as 3D reconstruction. As larger $\frac{B}{H}$ (i.e., 0.48) are not represented in the training data (see \cref{fig:bhtraining}), the performance of our models coupled with the MLP similarity deteriorates. PSMNet is slightly better compared to our best performing architecture MS-AFF+MLP. As the MLP clearly specializes to seen $\frac{B}{H}$, feature representations alone remain powerful and expressive: MS-AFF {cosine} outperforms PMSNet by 7.73 $\%$ on $D_{1}$. For $\frac{B}{H}=0.48$, MS-AFF with or without MLP is more robust to acquisition geometric changes than the rest of the models. When no particular $\frac{B}{H}$ configuration is privileged, cosine-based similarities are more advantageous in presence of varying $\frac{B}{H}$.\par
\vspace{-2pt}
On the unseen WV-3 stereo pairs, the DeepSim-Nets reconstruct buildings' details and boundaries more faithfully and with less regularization, {whereas PSMNet outputs fuzzy buildings and erases fine details (\cref{fig:epipsat_wv3}). Local methods (NCC, MC-CNN) produce noisy surfaces with disparity jumps on repetitive rooftop patterns.} 
On the unseen Pléiades 1B stereo pairs, we evaluated our models and compared them with the MC-CNN acrt trained using our sampling scheme. Although our best-performing model was \Unet32+MLP, it was outperformed by the MC-CNN acrt (see \cref{tab:errorrates}). Interestingly, deactivating the MLP improved our model's transferability and the accuracy of the disparity maps. In contrast to PSMNet, which merged buildings and their shadows into a single entity, our method accurately classified shadows as ground features. Notably, our \Unet32 {cosine} model showed a significant 3.19\% improvement in the $D_{1}$ metric when compared to MC-CNN acrt.
\section{Conclusion}
In this study, we have presented several variants of DeepSim-Nets for learning stereo-correspondence, which outperform standard hybrid methods on all examined metrics. Our networks can allocate sets of pixels in epipolar geometry and learn similarities simultaneously, overcoming the locality constraint and leveraging more global, context-aware, and transferable similarity cues. We have designed a sample mining scheme that improves deep feature extractors and enables occlusion detection in our models through contrastive training. Our flexible and lightweight MS-AFF model is designed to fit large multi-scale iterative dense matching pipelines, and generalizes well to unseen aerial and satellite stereo pairs.
\section{Acknowledgements}
This work was funded by Thales. We thank the AI4Geo project for the HPC resources and the Pléiades 1B dataset.
{\small
\bibliographystyle{ieee_fullname}
\bibliography{egbib}
}

\end{document}